\pdfoutput=1

\documentclass[11pt]{article}

\usepackage{acl}

\usepackage{times}
\usepackage{latexsym}
\usepackage{colortbl}

\usepackage[T1]{fontenc}

\usepackage[utf8]{inputenc}

\usepackage{microtype}

\usepackage{inconsolata}

\usepackage{graphicx}

\usepackage{multirow}
\usepackage{amsmath,amssymb,amsfonts}
\usepackage{amsthm}
\usepackage{mathrsfs}
\usepackage{xcolor}
\usepackage{float}
\usepackage{textcomp}
\usepackage{manyfoot}
\usepackage{booktabs}
\usepackage{algorithm}
\usepackage{algorithmicx}
\usepackage{algpseudocode}
\usepackage{listings}
\usepackage{courier}
\usepackage{placeins}
\usepackage{url}            
\usepackage{nicefrac}       
\usepackage{microtype}      
\usepackage[nolist]{acronym}
\usepackage[acronym]{glossaries}
\usepackage{siunitx}
\usepackage{longtable}
\usepackage{enumitem}       
\usepackage{wrapfig}        
\usepackage[font={small}]{caption}
\DeclareCaptionType{listing}[Listing][List of Listings]
\usepackage{comment}
\usepackage{framed}
\usepackage{xltabular}
\usepackage{chngcntr}
\usepackage{multicol}
\usepackage{subfigure}
\usepackage{lipsum}
\usepackage{hyperref}       

\usepackage{cleveref}       

\newcommand{\LegalSwissIcon}{\includegraphics[width=1.5em]{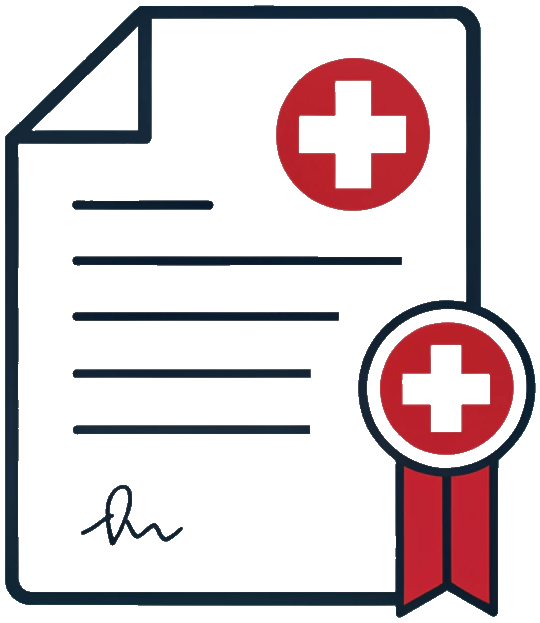}}

\title{
    \begin{minipage}{2em}
        \centering
        \LegalSwissIcon
    \end{minipage}
    \hspace{-5em}
    \begin{minipage}{0.85\textwidth}
        \centering
        \raisebox{-.5\height}{
        \hspace{4.5em}
        \shortstack{SwiLTra-Bench: The \textbf{Swi}ss \textbf{L}egal \textbf{Tra}nslation \textbf{Bench}mark}
        }
    \end{minipage}
}

\newcommand{\harvey}{\includegraphics[width=1.4em]{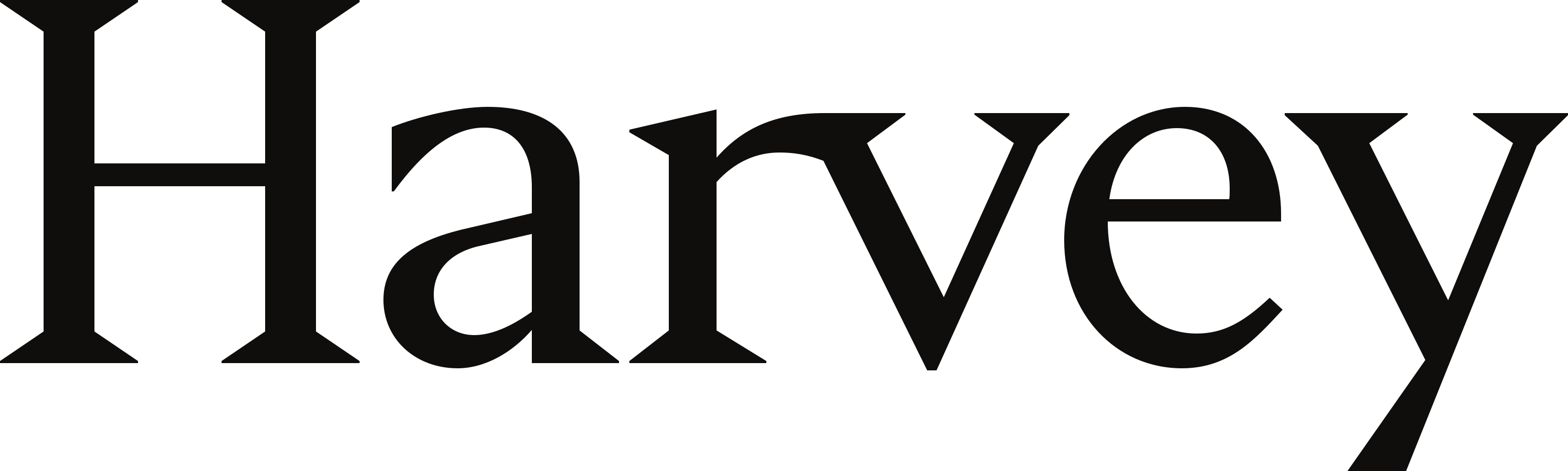}}
\newcommand{\ethlogo}{\includegraphics[width=1.2em]{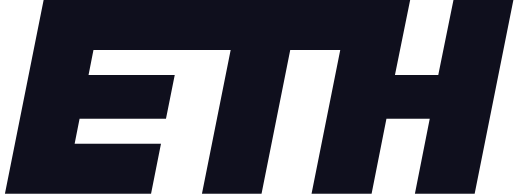}}
\newcommand{\court}{\includegraphics[width=1em]{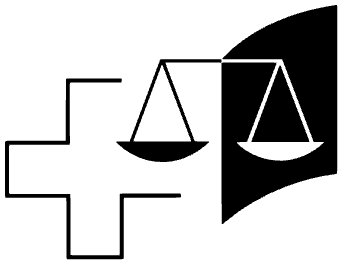}}
\newcommand{\uzh}{\includegraphics[width=1.1em]{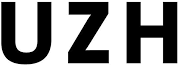}}
\newcommand{\basel}{\includegraphics[width=0.7em]{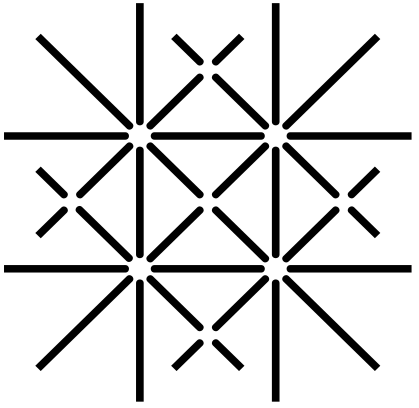}}
\newcommand{\genf}{\includegraphics[width=0.9em]{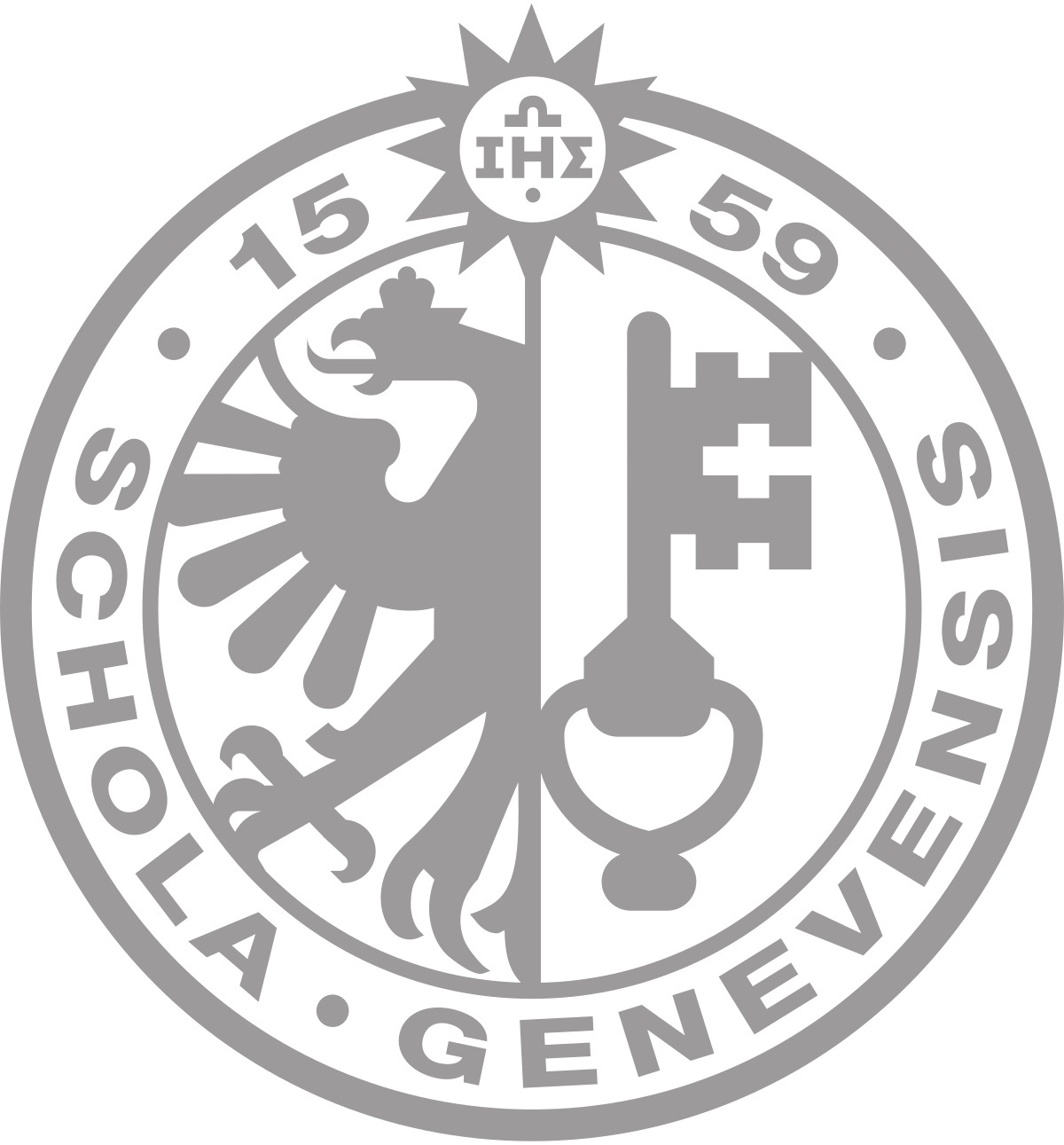}}
\newcommand{\lausanne}{\includegraphics[width=1.8em]{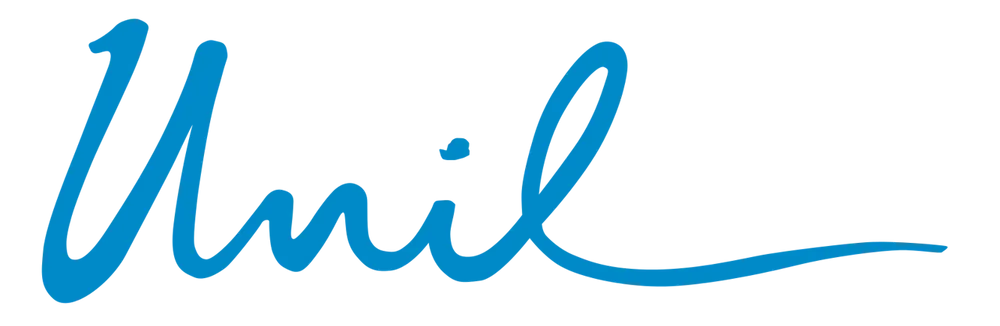}}
\newcommand{\solothurn}{\includegraphics[width=0.6em]{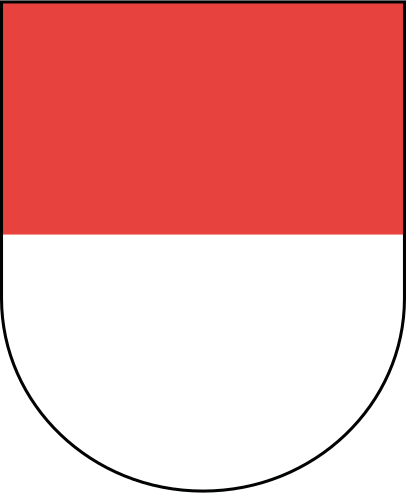}}
\newcommand{\mpi}{\includegraphics[width=0.8em]{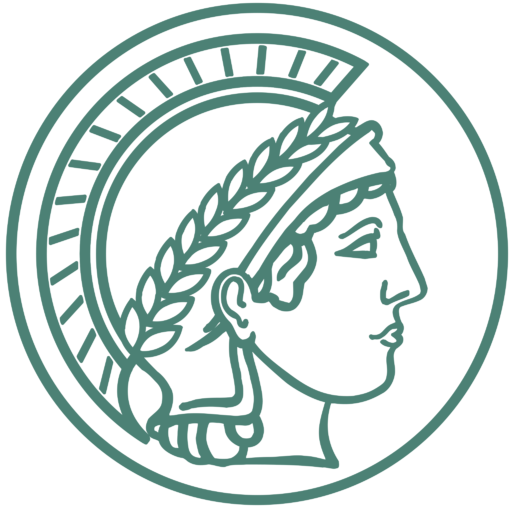}}

\author{
Joel Niklaus$^{\harvey}$\\
\textbf{
Jakob Merane$^{\ethlogo, \lausanne, \mpi}$\quad
Luka Nenadic$^{\ethlogo}$\quad
Sina Ahmadi$^{\uzh}$\quad
Yingqiang Gao$^{\uzh}$
}\\ \textbf{
Cyrill A. H. Chevalley$^{\basel}$\quad
Claude Humbel$^{\uzh}$\quad
Christophe Gösken$^{\ethlogo}$\quad
Lorenzo Tanzi$^{\genf}$\quad
}\\ \textbf{
Thomas Lüthi$^{\solothurn}$\quad
Stefan Palombo$^{\harvey}$\quad
Spencer Poff$^{\harvey}$\quad
Boling Yang$^{\harvey}$\quad
Nan Wu$^{\harvey}$
}\\ \textbf{
Matthew Guillod$^{\harvey}$\quad
Robin Mamié$^{\court}$\quad
Daniel Brunner$^{\court}$\quad
Julio Pereyra$^{\harvey}$\quad
Niko Grupen$^{\harvey}$
}\\
$^{\harvey}$Harvey\quad
$^{\ethlogo}$ETH Zurich\quad
$^{\court}$Swiss Federal Supreme Court\\
$^{\uzh}$University of Zurich\quad
$^{\basel}$University of Basel\quad
$^{\genf}$University of Geneva\quad
$^{\lausanne}$University of Lausanne\\
$^{\solothurn}$Canton of Solothurn\quad
$^{\mpi}$Max Planck Institute for Research on Collective Goods\\
\small{\textbf{Correspondence:} \href{mailto:joel@niklaus.ai}{joel@niklaus.ai}}
}

\begin{document}

\maketitle

\begin{acronym}[UMLX]
    \acro{FSCS}{Federal Supreme Court of Switzerland}
    \acro{SCI}{Supreme Court of India}
    \acro{ECHR}{European Convention of Human Rights}
    \acro{ECtHR}{European Court of Human Rights}
    \acro{SCOTUS}{Supreme Court of the United States}
    \acro{SPC}{Supreme People's Court of China}
    \acro{SJP}{Swiss-Judgment-Prediction}
    \acro{ASO}{Almost Stochastic Order}
    \acro{ILDC}{Indian Legal Documents Corpus}
    
    \acro{US}{United States}
    \acro{EU}{European Union}

    \acro{NLP}{Natural Language Processing}
    \acro{ML}{Machine Learning}
    \acro{LJP}{Legal Judgment Prediction}
    \acro{SJP}{Swiss-Judgment-Prediction}
    \acro{PJP}{Plea Judgment Prediction}
    
    \acro{BERT}{Bidirectional Encoder Representations from Transformers}
    \acro{LSTM}{ Long Short-Term Memory }
    \acro{GRU}{Gated Recurrent Unit}
    \acro{BiLSTM}{Bidirectional Long Short-Term Memory}
    \acro{CNN}{Convolutional Neural Networks}

    \acro{PLM}{pre-trained Language Model}
    \acro{LM}{Language Model}

    \acro{RTD}{Replaced Token Detection}

    \acro{CLT}{Cross-Lingual Transfer}
    \acro{HRL}{high resource language}
    \acro{LRL}{low resource language}

    \acro{POS}{Part-of-Speech}
    
    \acro{SLTC}{Single Label Text Classification}
    \acro{MLTC}{Multi Label Text Classification}
    \acro{TC}{Text Classification}
    \acro{NLU}{Natural Language Understanding}
    \acro{IR}{Information Retrieval}
    \acro{NER}{Named Entity Recognition}
    \acro{NLU}{Natural Language Understanding}
    \acro{QA}{Question Answering}
    \acro{NLI}{Natural Language Inference}

    \acro{GNB}{Gaussian Naive Bayes}
    \acro{DT}{Decision Tree}
    \acro{SVM}{Support-Vector Machine}
    \acro{RF}{ Random Forest}
    \acro{XGBoost}{eXtreme Gradient Boosting}
    \acro{MLIR}{Multilingual Information Retrieval}
    \acro{IR}{Information Retrieval}
    \acro{NDCG}{Normalized Discounted Cumulative Gain}
    \acro{LD}{Leading Decision}
    \acro{FSCD}{Federal Supreme Court Decisons}
    \acro{SFCS}{Swiss Federal Supreme Court}
    \acro{LLM}{Large Language Model}
    \acro{LM}{Language Model}

    \acro{CVG}{Court View Generation}
    \acro{JP}{Judgment Prediction}
    \acro{LAP}{Law Area Prediction}
    \acro{SLAP}{Sub Law Area Prediction}
    \acro{CP}{Criticality Prediction}
    \acro{LDS}{Leading Decision Summarization}
    \acro{CE}{Citation Extraction}
    
    \acro{SBERT}{Sentence-Bert}
    \acro{Doc2Doc}{Document-to-Document}
    \acro{mUSE}{Multilingual Universal Sentence Encoder}

\end{acronym}

\makeatletter
\setlength{\@fptop}{3em}
\makeatother

\vspace*{3em}

\begin{abstract}

In Switzerland legal translation is uniquely important due to the country's four official languages and requirements for multilingual legal documentation. However, this process traditionally relies on professionals who must be both legal experts and skilled translators---creating bottlenecks and impacting effective access to justice. To address this challenge, we introduce SwiLTra-Bench, a comprehensive multilingual benchmark of over 180K aligned Swiss legal translation pairs comprising laws, headnotes, and press releases across all Swiss languages along with English, designed to evaluate LLM-based translation systems. Our systematic evaluation reveals that frontier models achieve superior translation performance across all document types, while specialized translation systems excel specifically in laws but under-perform in headnotes. Through rigorous testing and human expert validation, we demonstrate that while fine-tuning open SLMs significantly improves their translation quality, they still lag behind the best zero-shot prompted frontier models such as Claude-3.5-Sonnet. Additionally, we present SwiLTra-Judge, a specialized LLM evaluation system that aligns best with human expert assessments.
\vspace{-0.6cm}
\noindent
\begin{center}
\begin{minipage}{\columnwidth}
\centering
\raisebox{-0.15cm}{\hspace{-4em} \includegraphics[height=0.5cm]{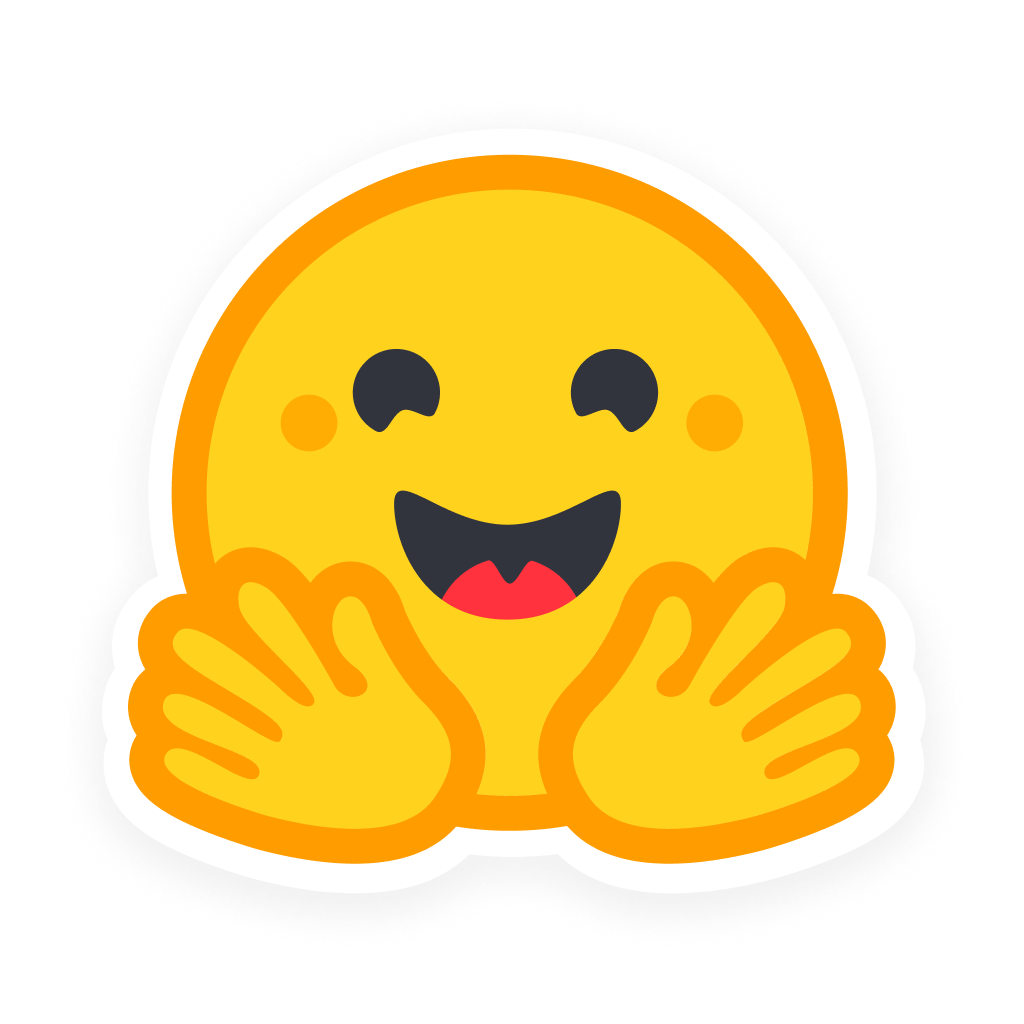}}~
\href{https://huggingface.co/collections/joelniklaus/swiltra-bench-67c569a2ada47e4549733deb}{\textbf{Datasets}} \quad
\raisebox{-0.15cm}{\includegraphics[height=0.5cm]{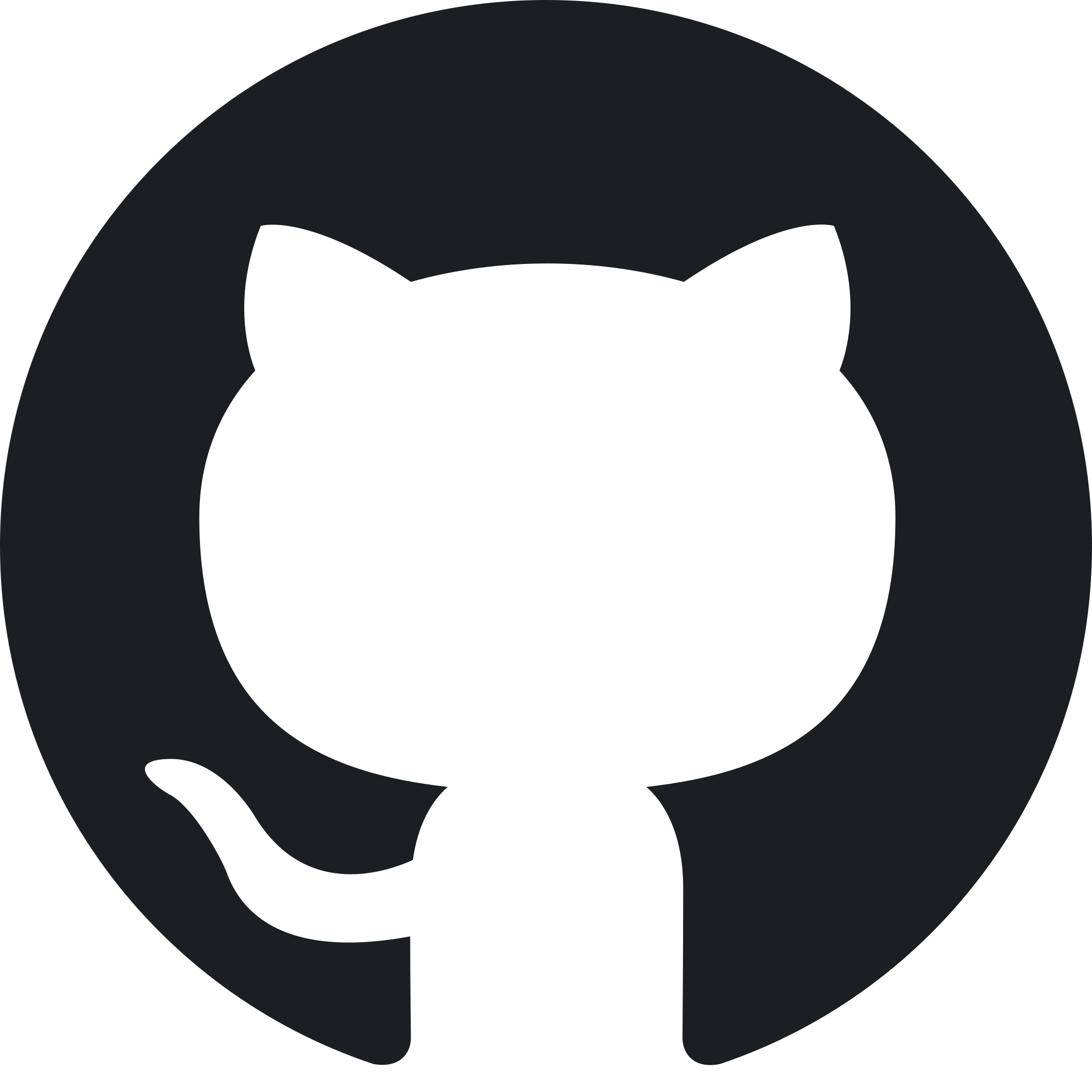}}~
\href{https://github.com/JoelNiklaus/SwissLegalTranslations}{\textbf{Code}}
\end{minipage}
\end{center}
\end{abstract}

\section{Introduction}

Neural Machine Translation (NMT) is one of the most studied Natural Language Processing (NLP) tasks. 
From encoder-decoder pipelines \citep{dai2015semi, vaswani2017attention} to modern decoder-only models \citep{brown2020language, touvron2023llama} NMT systems based on large language models (LLMs) have in recent years achieved notable advancements in translating texts across various genres \citep{ou2023songs, zhang2024good, han2024neural} and in both high- and low-resource languages \citep{moslem2023adaptive, vilar2023prompting, alves2023steering, oliver2024training}. Nevertheless, the shortage of high-quality multilingual parallel legal translation data for training LLMs has hindered the performance of state-of-the-art NMT systems in translating legal texts. This limitation is primarily due to the discourse structures \citep{wiesmann2019machine} and specialized terminology \citep{katz2023natural} of legal texts, which consequently result in the current limited degree of automation for translation in the legal domain.

\begin{figure}[t]
    \centering
    \vspace{5em}
    \includegraphics[width=\columnwidth]{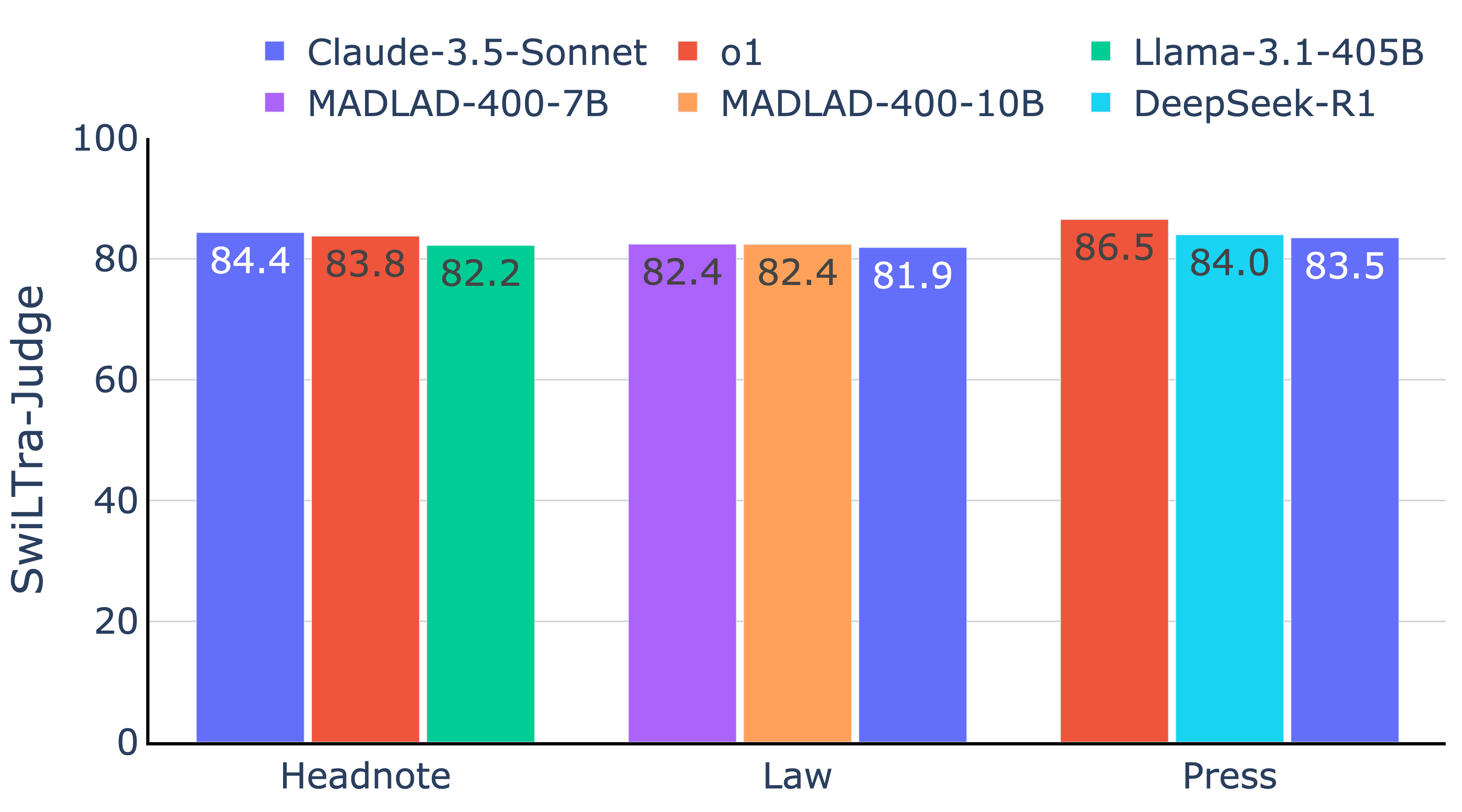}
    \caption{Best models per task.}
    \label{fig:best-per-task}
\end{figure}

In multilingual countries like Switzerland, where legal documents are primarily translated manually by experts, developing reliable NMT systems for legal texts would significantly improve governmental efficiency and reduce administrative bottlenecks \citep{martinez2020customized}. Beyond operational benefits, such systems could democratize access to legal information by enabling faster and more cost-effective translations across multiple national languages. Especially in lower-resourced languages like Romansh where full translation coverage is not currently economical, support from high-quality NMT systems could be game-changing. This broader accessibility would enhance the transparency of political decision-making and promote more inclusive civic participation \citep{moniz2023towards}. The potential impact extends beyond government operations to the private sector where law firms and businesses operating across linguistic regions could benefit from improved legal translation capabilities, potentially reducing costs and accelerating legal processes while maintaining accuracy. Although initial efforts have been made to develop NMT systems for translating Swiss legal documents \citep{martinez2020customized, canavese2024translators}, it remains unclear how well current LLMs perform on large benchmarks for translating Swiss legal texts, both in zero-shot and fine-tuning settings.

To address the shortage of Swiss legal training data and advance legal translation, we present three main contributions:
\begin{enumerate}[noitemsep,nolistsep,leftmargin=*]
    \item \textsc{SwiLTra-Bench}: A large-scale benchmark of over 180K aligned Swiss legal translation pairs (laws, court decisions, press releases) spanning five languages (the four official Swiss languages plus English), substantially expanding available training data.
    \item Comprehensive Model Comparison: The first large-scale evaluation of frontier LLMs and fine-tuned open SLMs on Swiss legal translations in both zero-shot and fine-tuning settings, providing insights into their relative strengths.
    \item \textsc{SwiLTra-Judge}: An LLM-based method aligned with human expert annotations, offering a reliable automated framework to assess translation quality.
\end{enumerate}


Our main findings are: a) frontier models consistently perform well across translation tasks; b) translation specific systems like MADLAD-400 are strong on laws but fall behind on headnotes; c) fine-tuning open SLMs drastically improves their translation quality but they are still behind zero-shot prompted frontier models; d) translation quality is uniform across languages; and e) agreement among human experts is higher for law translations than for headnotes.


\section{SwiLTra-Bench} 

To support research in NMT, text alignment, and legal document processing, we present SwiLTra-Bench---a dataset uniting original legal texts and press releases on key court rulings.

\begin{table}[htb]
    \centering
    \subtable[CH-Law-Trans dataset.]{
    \resizebox{\columnwidth}{!}{    
    \begin{tabular}{clcccccc}
    \toprule
     Source & Split & \#file & \#de & \#fr & \#it & \#rm & \#en \\ 
     \midrule
    \multirow{3}{*}{Law} & Train & 5,206 & 5,206 & 5,206 & 5,206 & 51 & 219 \\
    & Valid & 10 & 10 & 10 & 10 & 10 & 10 \\
    & Test & 20 & 20 & 20 & 20 & 20 & 20 \\
    \midrule
    \multirow{3}{*}{Article} & Train & 129,070 & 126,308 & 127,049 & 126,223 & 8,680 & 16,347 \\
    & Valid & 789 & 785 & 785 & 784 & 785 & 785 \\
    & Test & 740 & 738 & 738 & 738 & 738 & 738 \\
    \midrule
    \multirow{3}{*}{Paragraph} & Train & 153,970 & 145,106 & 146,953 & 145,267 & 19,556 & 32,499 \\
    & Valid & 1,490 & 1,441 & 1,438 & 1,437 & 1,441 & 1,439 \\
    & Test & 1,214 & 1,176 & 1,178 & 1,178 & 1,177 & 1,176 \\
    \bottomrule
    \end{tabular}
    }
    \label{tab:law_trans}
    }
    \centering
    \subtable[CH-Headnote-Trans dataset.]{
    \resizebox{0.7\columnwidth}{!}{
    \centering
    \begin{tabular}{clcccc}
    \toprule
     Source & Split & \#file & \#de & \#fr & \#it  \\ 
     \midrule
    \multirow{3}{*}{BGE} & Train & 13,330 & 13,330 & 13,330 & 13,330  \\
    & Valid & 1,900 & 1,900 & 1,900 & 1,900  \\
    & Test & 3,801 & 3,801 & 3,801 & 3,801  \\
    \midrule
    \multirow{3}{*}{Regest} & Train & 13,550 & 13,550 & 13,550 & 13,550  \\
    & Valid & 1,924 & 1,924 & 1,924 & 1,924 \\
    & Test & 3,890 & 3,890 & 3,890 & 3,890  \\
    \midrule
    \multirow{3}{*}{Text} & Train & 26,008 & 26,008 & 26,008 & 26,008  \\
    & Valid & 3,805 & 3,805 & 3,805 & 3,805 \\
    & Test & 7,316 & 7,316 & 7,316 & 7,316  \\
    \bottomrule
    \end{tabular}
    }
    \label{tab:bge_trans}
    }
    \centering
    \subtable[CH-Press-Trans dataset.]{
    \resizebox{0.7\columnwidth}{!}{
    \begin{tabular}{clcccc}
    \toprule
    Source & Split & \#file & \#de & \#fr & \#it \\
    \midrule
    \multirow{3}{*}{Press} & Train & 867 & 867 & 867 & 152 \\
    & Valid & 100 & 100 & 100 & 100 \\
    & Test & 200 & 200 & 200 & 200 \\ 
    \bottomrule
    \end{tabular}
    }
    \label{tab:press_trans}
    }
    \caption{Overall \textsc{SwiLTra-Bench} corpus statistics. \#file indicates the total number of files collected, while \#de, \#fr, \#it, \#rm, and \#en represent the ones in the respective languages.}
    \label{tab:dataset-overview}
\end{table}

\subsection{Data Collection}

SwiLTra-Bench contains three sub-datasets: 
\begin{enumerate}[noitemsep,nolistsep,leftmargin=*]
    \item Swiss Law Translations (CH-Law-Trans), including law-level (entire legal documents), article-level (individual articles), and paragraph-level (paragraphs within articles) translations.
    \item Headnote Translations (CH-Headnote-Trans) of Swiss Supreme Court landmark court decisions (\textit{``Bundesgerichtsentscheide'' (BGE)} in German, \textit{``Arrêts du Tribunal fédéral'' (ATF)} in French, and \textit{``Decisioni principali del Tribunale federale svizzero'' (DFT)} in Italian) at the BGE-level (complete summaries of court decisions), regest-level (summaries focused on core legal issues), and text-level (detailed extraction of specific legal statements).
    \item Swiss Supreme Court Press Release Translations (CH-Press-Trans).
\end{enumerate}

All datasets contain parallel translations in German (de), French (fr), and Italian (it). Additionally, for CH-Law-Trans, some documents contain translations in Romansh (rm) and English (en).

We provide details of the data structure with concrete dataset examples in \autoref{sec:corpus-examples}. 

\subsection{Dataset Splits}

We first segment each dataset by a unique identifier (entire laws, entire headnotes, and entire press releases) to ensure that no single law, headnote, or press release is split across training, validation, and test. For laws, we prioritize examples for the validation and test splits that (1) have more language versions (to guarantee good multilingual coverage), (2) have an official abbreviation (since abbreviations are only set for those laws that are presumed to be cited frequently\footnote{\href{https://www.bk.admin.ch/apps/gtr/de/index.html}{https://www.bk.admin.ch/apps/gtr/de/index.html}}, which we consider a good proxy for practical importance, (3) have shorter text lengths to make evaluation faster and cheaper, and (4) have newer applicability dates so that more recent and multilingual laws are prioritized for validation and testing, resulting in a more realistic evaluation setting. For headnotes, we similarly prioritize those with more recent publication years for validation and test. Finally, for press releases, we focus on maximizing multilingual coverage by ensuring all validation and test examples are available in all present languages (German, French and Italian). The training sets contain all examples not held out for validation or testing.

\subsection{Data Statistics}

\autoref{tab:dataset-overview} presents the overall statistics of the three datasets included in SwiLTRa-Bench. We visualize the training set text lengths for the shortest levels used for training and evaluation in \autoref{fig:word-count}. For completeness, we show histograms for all levels in \autoref{fig:word-count-headnote-press-app} and \autoref{fig:word-count-law-app}. To calculate these statistics, we used an NLTK\footnote{\href{https://www.nltk.org}{https://www.nltk.org}} word tokenizer, splitting sentences based on whitespace and punctuation.

Existing parallel legal corpora use automated methods for sentence alignment \cite{koehn2005europarl, un-parallel-corpus}. In SwiLTRa-Bench, we rely on the structure provided by the official government bodies such as law paragraphs embedded in the HTML, resulting in high-quality alignment.

\section{Experimental Setup}

\subsection{Evaluation}

To paint a representative picture of translation capabilities, we evaluate models across five main classes: 1) translation models, i.e., models specifically trained for translation tasks, 2) frontier models, i.e., large foundation models pre-trained on web-scale data and post-trained on diverse tasks, 3) reasoning models, i.e., models using significant resources at test time to improve output quality, 4) open models, i.e., typically small language models (SLMs) with publicly available weights, and 5) fine-tuned models, i.e., models specifically fine-tuned on SwiLTRa-Bench.
We conducted our evaluation using the \texttt{lighteval} framework due to its ease of use and good support for custom metrics.\footnote{\href{https://github.com/huggingface/lighteval}{https://github.com/huggingface/lighteval}}

\subsubsection{Metrics}


We evaluated translations using lexical (BLEU \cite{papineni_bleu_2002}, ChrF \cite{popovic-2015-chrf}, METEOR \cite{banerjee_meteor_2005}) and model-based metrics (BERTScore \cite{zhang_bertscore_2020}, BLEURT \cite{sellam-etal-2020-bleurt}, XCOMET \cite{guerreiro-xcomet}, GEMBA-MQM \cite{kocmi-federmann-2023-gemba}). Due to the 512-token limit, BLEURT and XCOMET cannot process press releases. Given GEMBA-MQM's strong correlation with human judgments, we prioritized it alongside XCOMET, METEOR, and ChrF, ensuring both lexical and trained metrics for diversity.


\subsection{Fine-tuning}

To provide an overview of the current open SLM landscape, we fine-tuned Gemma-2 2B and 9B \cite{gemma2shortened}, Llama 1B, 3B and 8B \cite{llama3shortened}, 
Phi-3.5 mini and Phi-3 medium \cite{phi3shortened}, and Qwen2.5 0.5B, 1.5B, 3B, 7B, 14B and 32B \cite{qwen2.5} models on our dataset. 
We fine-tuned using Hugging Face \texttt{transformers}\footnote{\href{https://github.com/huggingface/transformers}{https://github.com/huggingface/transformers}} and \texttt{unsloth}\footnote{\href{https://github.com/unslothai/unsloth}{https://github.com/unslothai/unsloth}} using 4-bit quantization and 8bit AdamW \cite{loshchilov2019decoupledweightdecayregularization, dettmers20228bitoptimizersblockwisequantization} on a single 80GB NVIDIA H100 GPU. We used rank stabilized LoRA \cite{hu_lora_2021, kalajdzievski2023rankstabilizationscalingfactor} with rank 16 and alpha 16. We trained with the model's native chat template on sequence length 512, covering more than 99\% of the training dataset and truncating the rest. The instruction template was simply:
\begin{quote}
\begin{tabular}{@{}l@{ }l}
\textbf{\{source language\}:} & \{source text\}\\
\textbf{\{target language\}:} & \{target text\}
\end{tabular}
\end{quote}
We trained on the entire training set for maximal coverage of the data. We used packing, weight decay 0.01, batch size 128 and early stopping with patience 3. In most cases, the lowest evaluation loss is reached after exactly 1 epoch. We used a linear learning rate schedule with 1000 warmup steps and learning rate $1e-4$. 
We manually tuned the learning rate ($1e-5$ - $1e-3$), weight decay (0.01, 0.1), label smoothing (factor 0, 0.01, 0.1) and LoRA rank (16, 128).
We used the train and validation sets of the Law and Headnote translations on the lowest (shortest) levels, i.e., the paragraph and text levels. 
For all fine-tuned models, we used the instruction-tuned variant since they have shown to better adapt to new tasks \cite{niklaus_flawn-t5_2024}.
\section{Results and Analysis}

In the tables, we bolded the highest and underlined the second highest score per metric. Unless stated otherwise, we excluded Romansh from the evaluations to ensure comparability, since it is not supported by the translation models. Unless stated otherwise, results are averaged over source languages, target languages, and tasks. In general, we considered the law and headnote translation tasks at the highest granularity (paragraph-level and text-level) so we can compare all model categories (translation models and fine-tuned models are optimized for shorter sequence lengths). We show all metrics with standard errors obtained through bootstrapping. Higher values are better for all metrics. 


\subsection{Statistical Significance Testing}
The model comparisons presented throughout this section are descriptive in nature. When we state that one model ``outperforms'' or ``outcompetes'' another, we are describing observed differences in the reported metrics; we do not claim statistical significance unless explicitly stated. We only apply null hypothesis significance testing (NHST) in those instances where we explicitly report statistically significant results. In such cases, we provide information about the tests used, statistics obtained, and effect sizes.

\subsection{Translation Models}

We compare translation models in \autoref{tab:translation-models}. Surprisingly, Google-Translate performs poorly compared to open translation models like MADLAD-400 \cite{madlad400} and Tower-Instruct. Facebook's SeamlessM4T \cite{seamlessm4t} model's text-to-text capabilities also underwhelm. MADLAD-400 performs very well, outperforming GPT-4o on XCOMET. The Tower \cite{tower} models land somewhere in between.

\begin{table}[!htb]
\resizebox{\columnwidth}{!}{\begin{tabular}{llrrrr}
\toprule
\textbf{Model} & \textbf{Size} & \textbf{↑ GEMBA-MQM} & \textbf{↑ XCOMET} & \textbf{↑ METEOR} & \textbf{↑ ChrF}\\
\midrule
Google-Translate & N/A & \cellcolor{teal!57}53.20 ± 0.2 & \cellcolor{teal!44}64.61 ± 0.1 & \cellcolor{teal!47}41.15 ± 0.1 & \cellcolor{teal!61}47.81 ± 0.1 \\
MADLAD-400-3B & 3B & \textbf{\cellcolor{teal!70}62.89 ± 0.1} & \underline{\cellcolor{teal!69}86.82 ± 0.1} & \cellcolor{teal!57}42.44 ± 0.1 & \cellcolor{teal!68}51.36 ± 0.1 \\
MADLAD-400-7B & 7B & \underline{\cellcolor{teal!69}62.66 ± 0.1} & \textbf{\cellcolor{teal!70}87.40 ± 0.1} & \underline{\cellcolor{teal!67}43.70 ± 0.1} & \underline{\cellcolor{teal!68}51.67 ± 0.1} \\
MADLAD-400-10B & 10B & \cellcolor{teal!68}61.46 ± 0.1 & \cellcolor{teal!69}86.65 ± 0.1 & \cellcolor{teal!63}43.10 ± 0.1 & \textbf{\cellcolor{teal!70}52.24 ± 0.1} \\
SeamlessM4T & 2B & \cellcolor{teal!20}23.35 ± 0.2 & \cellcolor{teal!20}43.03 ± 0.1 & \cellcolor{teal!20}37.81 ± 0.1 & \cellcolor{teal!20}24.90 ± 0.1 \\
TowerInstruct-7B & 7B & \cellcolor{teal!58}54.04 ± 0.2 & \cellcolor{teal!53}72.97 ± 0.1 & \cellcolor{teal!51}41.65 ± 0.2 & \cellcolor{teal!53}43.00 ± 0.1 \\
TowerInstruct-13B & 13B & \cellcolor{teal!63}57.38 ± 0.2 & \cellcolor{teal!57}75.94 ± 0.1 & \textbf{\cellcolor{teal!70}43.95 ± 0.2} & \cellcolor{teal!63}48.46 ± 0.1 \\
\bottomrule
\end{tabular}}
\caption{Translation models across different families and sizes.}
\label{tab:translation-models}
\end{table}

\subsection{Frontier Models}

We show results for frontier and reasoning models in \autoref{tab:frontier-models}. GPT-4o underperforms both of its peers Claude-3.5-Sonnet and Llama-3.1-405B. This is particularly unexpected, as models tend to favor their own completions \cite{panickssery2024llmevaluatorsrecognizefavor}, and GEMBA-MQM is operated by GPT-4o. Claude-3.5-Sonnet demonstrates strong performance, competing closely with o1, the top-performing model. Surprisingly, o1-mini performs only on par with the other models at the smaller scale and even underperforms Claude-3.5-Haiku. Overall, Anthropic's models are really strong, and even more so from a cost-to-performance perspective compared to reasoning models like o1.

\begin{table}[!b]
\resizebox{\columnwidth}{!}{\begin{tabular}{llrrrr}
\toprule
\textbf{Model} & \textbf{Size} & \textbf{↑ GEMBA-MQM} & \textbf{↑ XCOMET} & \textbf{↑ METEOR} & \textbf{↑ ChrF}\\
\midrule
Claude-3.5-Sonnet & large & \cellcolor{teal!37}80.66 ± 0.2 & \underline{\cellcolor{teal!66}90.70 ± 0.1} & \cellcolor{teal!53}56.71 ± 0.2 & \cellcolor{teal!50}65.87 ± 0.1 \\
DeepSeek-V3 & large & \cellcolor{teal!33}80.04 ± 0.2 & \cellcolor{teal!62}89.77 ± 0.1 & \cellcolor{teal!52}56.60 ± 0.1 & \cellcolor{teal!69}69.99 ± 0.1 \\
DeepSeek-R1 & large & \cellcolor{teal!20}77.90 ± 0.2 & \cellcolor{teal!36}84.36 ± 0.1 & \cellcolor{teal!46}55.79 ± 0.1 & \cellcolor{teal!65}69.12 ± 0.1 \\
GPT-4o & large & \cellcolor{teal!34}80.27 ± 0.2 & \cellcolor{teal!20}80.96 ± 0.1 & \cellcolor{teal!45}55.56 ± 0.1 & \cellcolor{teal!38}63.27 ± 0.1 \\
Gemini-1.5-Pro & large & \cellcolor{teal!45}\cellcolor{teal!45}81.88 ± 0.2 & \cellcolor{teal!49}87.13 ± 0.1 & \underline{\cellcolor{teal!62}57.92 ± 0.1} & \underline{\cellcolor{teal!69}70.07 ± 0.1} \\
Llama-3.1-405B & large & \cellcolor{teal!43}81.59 ± 0.1 & \cellcolor{teal!60}89.37 ± 0.1 & \cellcolor{teal!37}54.48 ± 0.1 & \cellcolor{teal!60}68.07 ± 0.1 \\
Mistral-Large & large & \cellcolor{teal!45}\cellcolor{teal!45}81.88 ± 0.2 & \cellcolor{teal!49}\cellcolor{teal!49}87.04 ± 0.1 & \cellcolor{teal!40}54.86 ± 0.1 & \cellcolor{teal!40}63.71 ± 0.1 \\
o1 & large & \textbf{\cellcolor{teal!70}85.81 ± 0.1} & \textbf{\cellcolor{teal!70}91.35 ± 0.1} & \textbf{\cellcolor{teal!70}58.91 ± 0.1} & \textbf{\cellcolor{teal!70}70.11 ± 0.1} \\
Claude-3.5-Haiku & small & \cellcolor{teal!35}80.40 ± 0.2 & \cellcolor{teal!57}88.84 ± 0.1 & \cellcolor{teal!20}52.15 ± 0.2 & \cellcolor{teal!28}61.09 ± 0.1 \\
GPT-4o-mini & small & \underline{\cellcolor{teal!49}82.59 ± 0.2} & \cellcolor{teal!53}87.90 ± 0.1 & \cellcolor{teal!33}54.03 ± 0.1 & \cellcolor{teal!22}59.86 ± 0.1 \\
Gemini-1.5-Flash & small & \cellcolor{teal!38}80.76 ± 0.2 & \cellcolor{teal!41}85.33 ± 0.1 & \cellcolor{teal!43}55.35 ± 0.1 & \cellcolor{teal!48}65.44 ± 0.1 \\
Llama-3.3-70B & small & \cellcolor{teal!28}79.25 ± 0.2 & \cellcolor{teal!53}88.02 ± 0.1 & \cellcolor{teal!29}53.43 ± 0.1 & \cellcolor{teal!50}65.92 ± 0.1 \\
Mistral-Small & small & \cellcolor{teal!43}81.69 ± 0.2 & \cellcolor{teal!49}\cellcolor{teal!49}87.04 ± 0.1 & \cellcolor{teal!39}54.83 ± 0.1 & \cellcolor{teal!40}63.66 ± 0.1 \\
o1-mini & small & \cellcolor{teal!45}81.96 ± 0.2 & \cellcolor{teal!51}87.46 ± 0.1 & \cellcolor{teal!28}53.34 ± 0.1 & \cellcolor{teal!20}59.32 ± 0.1 \\
\bottomrule
\end{tabular}}
\caption{Frontier models across different families and sizes.}
\label{tab:frontier-models}
\end{table}

\subsection{Fine-tuned Models}

\begin{figure}[!t]
    \centering
    \includegraphics[width=0.87\columnwidth]{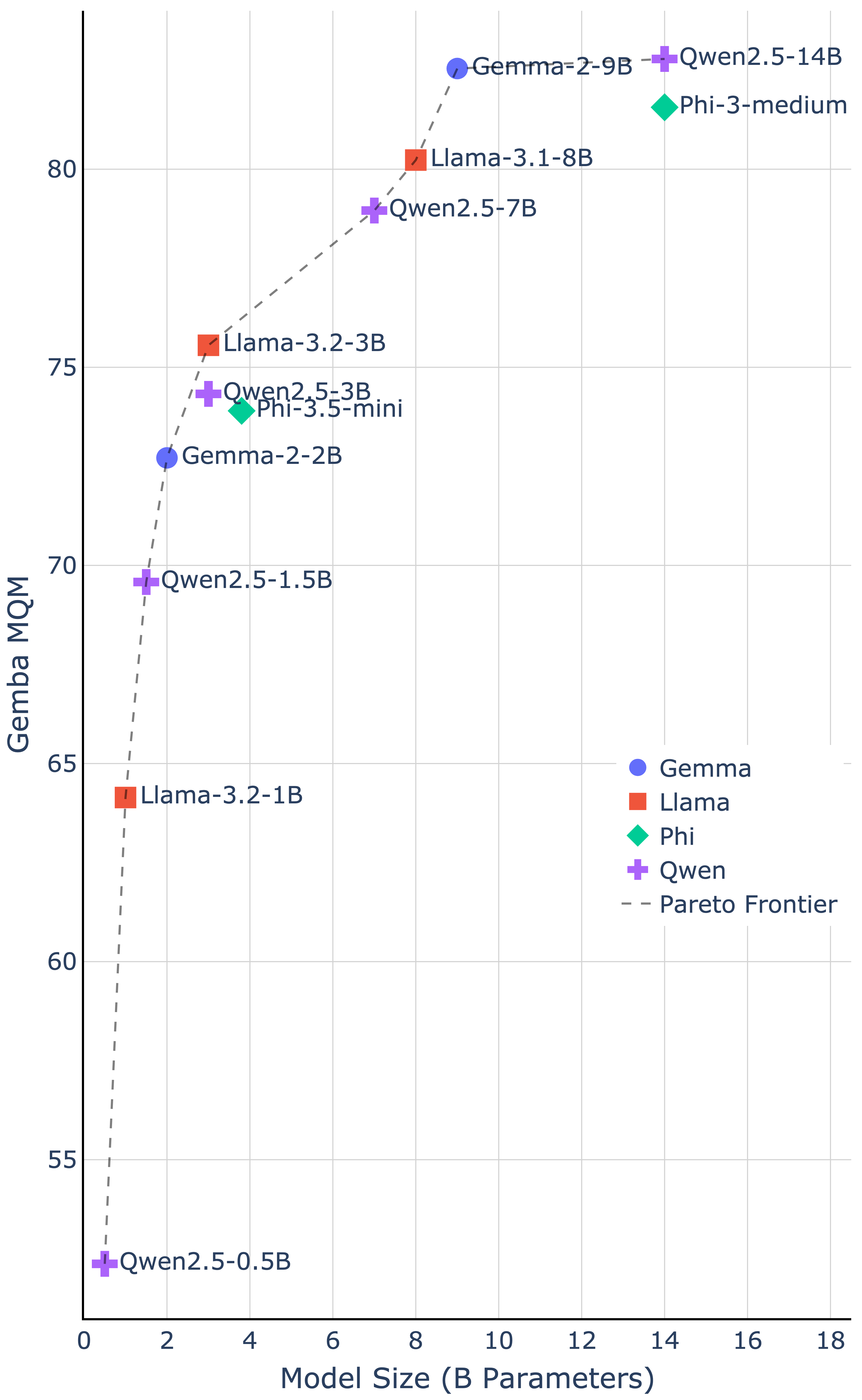}
    \caption{Finetuned models across different sizes.}
    \label{fig:finetuned-models}
\end{figure}

Fine-tuning leads to notable performance gains (see Appendix \autoref{tab:finetune-vs-base}). \autoref{fig:finetuned-models} presents fine-tuned models' performance across various sizes. The two Gemma models and particularly the Llama 1B and 3B models, advance the Pareto frontier, though performance starts to flatten at the 3B scale and plateaus after 9B parameters. Interestingly, both Phi models clearly underperform their peers.

\subsection{Performance Progression by Model Size}
\begin{figure}[htb]
    \centering
    \includegraphics[width=\columnwidth]{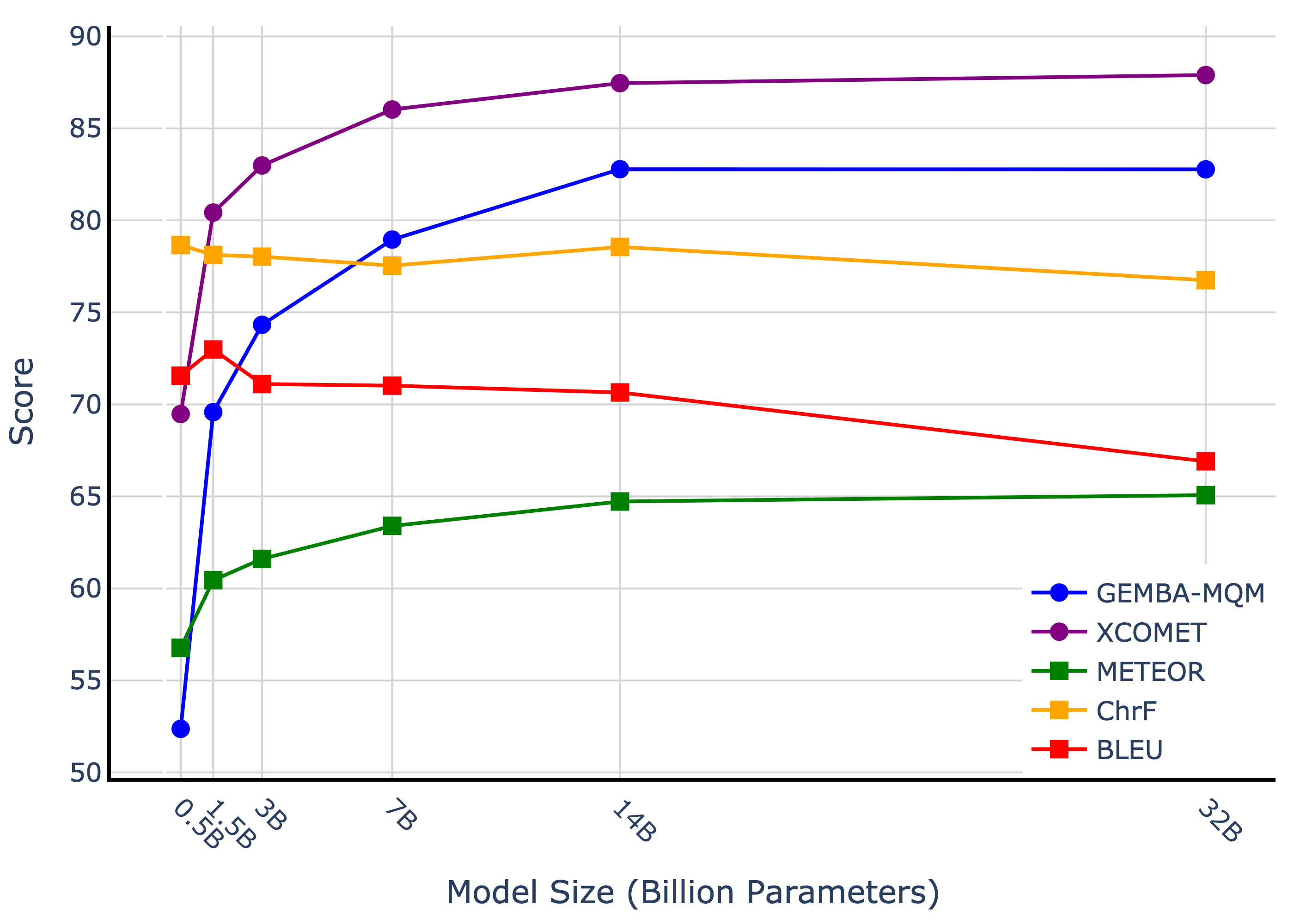}
    \caption{Lexical (square) and model-based (circle) metrics vs model size for finetuned Qwen models. }
    \label{fig:qwen-size-progression}
\end{figure}

The Qwen2.5 model family, with six sizes from 0.5B to 32B parameters, is ideal for studying performance progression over model size. We analyzed fine-tuned Qwen models up to 32B using five metrics (two model-based, three lexical) in \autoref{fig:qwen-size-progression}. 
METEOR is the only lexical metric well correlated with XCOMET and GEMBA-MQM. All three confirm a clear trend that larger models produce higher-quality translations. GEMBA-MQM shows the largest score range (GEMBA-MQM: 52.4 - 82.8 vs XCOMET: 69.5 - 87.9 and METEOR: 56.8 - 65.1) and making it most useful for differentiating models.
Interestingly, both ChrF and BLEU are negatively correlated with the model-based metrics on this task for the fine-tuned Qwen models. Beyond the inherent subjectivity in assessing translation quality, this may hint at the greater importance of a legal text's conveyed meaning over the mere use of certain exact terms. 

\subsection{Comparison Across Tasks}

In \autoref{tab:best-across-tasks}, we show the best models’ performance per category across tasks. The best open small model falls far behind the others but, with fine-tuning, overtakes the best translation model. It matches the smaller frontier models but still lags behind the larger ones. All models except MADLAD-400-7B perform better on headnote than law translation. While Sonnet competes with o1 on headnote and law translation, it falls off on press releases.

\begin{table}[!tb]
\resizebox{\columnwidth}{!}{\begin{tabular}{lllrrr}
\toprule
\textbf{Model} & \textbf{Category} & \textbf{Task} & \textbf{↑ GEMBA-MQM} & \textbf{↑ METEOR} & \textbf{↑ ChrF}\\
\midrule
o1 & reasoning & Headnote & \textbf{\cellcolor{teal!70}93.50 ± 0.1} & \cellcolor{teal!61}60.89 ± 0.1 & \cellcolor{teal!40}62.62 ± 0.2 \\
o1 & reasoning & Law & \underline{\cellcolor{teal!66}91.11 ± 0.1} & \cellcolor{teal!53}55.87 ± 0.1 & \cellcolor{teal!48}66.84 ± 0.1 \\
o1 & reasoning & Press & \cellcolor{teal!25}64.62 ± 0.4 & \cellcolor{teal!58}59.28 ± 0.3 & \textbf{\cellcolor{teal!70}78.38 ± 0.1} \\
Claude-3.5-Sonnet & frontier & Headnote & \cellcolor{teal!62}88.65 ± 0.1 & \cellcolor{teal!61}61.39 ± 0.1 & \cellcolor{teal!42}63.96 ± 0.2 \\
Claude-3.5-Sonnet & frontier & Law & \cellcolor{teal!58}85.71 ± 0.1 & \cellcolor{teal!47}52.16 ± 0.1 & \cellcolor{teal!60}73.15 ± 0.1 \\
Claude-3.5-Sonnet & frontier & Press & \cellcolor{teal!20}60.83 ± 0.8 & \cellcolor{teal!52}55.29 ± 0.5 & \cellcolor{teal!26}55.47 ± 0.1 \\
MADLAD-400-7B & translation & Headnote & \cellcolor{teal!50}80.54 ± 0.2 & \cellcolor{teal!56}57.71 ± 0.1 & \cellcolor{teal!49}67.49 ± 0.2 \\
MADLAD-400-7B & translation & Law & \cellcolor{teal!57}85.06 ± 0.2 & \cellcolor{teal!55}57.09 ± 0.2 & \cellcolor{teal!38}61.86 ± 0.2 \\
SLT-Qwen2.5-32B & finetuned & Headnote & \cellcolor{teal!53}82.58 ± 0.1 & \textbf{\cellcolor{teal!70}66.56 ± 0.1} & \cellcolor{teal!63}75.17 ± 0.2 \\
SLT-Qwen2.5-32B & finetuned & Law & \cellcolor{teal!50}80.80 ± 0.2 & \underline{\cellcolor{teal!66}64.41 ± 0.1} & \underline{\cellcolor{teal!67}76.90 ± 0.1} \\
Qwen2.5-14B & open & Headnote & \cellcolor{teal!33}69.88 ± 0.2 & \cellcolor{teal!39}47.09 ± 0.1 & \cellcolor{teal!22}53.58 ± 0.2 \\
Qwen2.5-14B & open & Law & \cellcolor{teal!23}63.04 ± 0.2 & \cellcolor{teal!20}34.33 ± 0.1 & \cellcolor{teal!20}52.02 ± 0.1 \\
\bottomrule
\end{tabular}}
\caption{Best models per category across different tasks.}
\label{tab:best-across-tasks}
\end{table}

\subsection{Comparison Across Languages}

\begin{figure*}
    \centering
    \includegraphics[width=\textwidth]{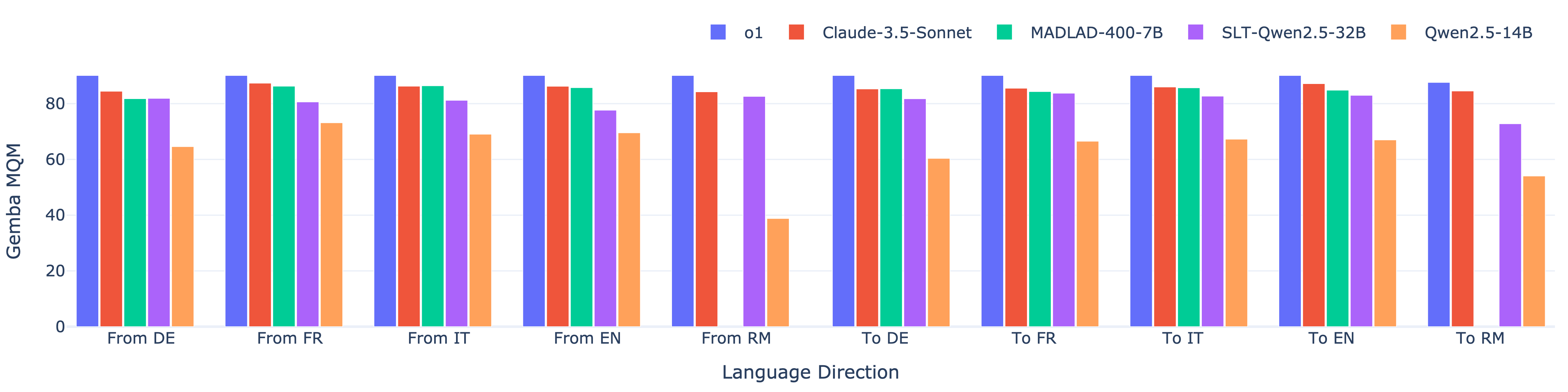}
    \caption{Best models per category across languages.}
    \label{fig:language-comparison}
\end{figure*}

In \autoref{fig:language-comparison}, we compare the best models per category across language directions on CH-Law-Trans. Performance to and from German, French, and Italian is homogeneous across models. When translating from English to the other languages, all models perform worse than from the three main Swiss languages. Since the English source texts are already translations and are not legally binding, the federal translators may have applied less rigor in generating them, potentially resulting in lower quality and slight deviations. Anecdotally, the lawyers co-authoring this work confirm that the English source texts are occasionally less precise. So, the lower scores may also indicate that the judge model bases its grading on imperfect source text. 

Romansh is a low-resource language and only spoken by less than 50K people in Switzerland.\footnote{\href{https://www.bfs.admin.ch/bfs/de/home/statistiken/bevoelkerung/sprachen-religionen/sprachen.html}{https://www.bfs.admin.ch/bfs/de/home/statistiken/\allowbreak bevoelkerung/sprachen-religionen/sprachen.html}}  As our dataset consists of the entire data readily available for federal laws, Supreme Court headnotes, and Supreme Court press releases in Switzerland, it is not possible to extend the Romansh coverage. Adding additional data sources is not a viable option due to likely lower data quality and limited sentence level alignment across languages. Romansh is not supported by most translation models such as MADLAD-400. In our opinion, the fact that these models don't support Romansh at the moment highlights the value of our dataset for low-resource languages. Indeed, our dataset now enables teams building translation models to train and evaluate on >160K and >8K Romansh translation pairs respectively. Surprisingly, o1 and Sonnet still perform very well when translating from Romansh to other languages. When translating to Romansh, all models' quality drops off, sometimes sharply. Perhaps similarly to humans, also for LLMs speaking or writing a language seems harder than understanding it.


\section{Expert Evaluation}
\label{sec:expert-evaluation}

To study how well human legal experts agree with the automated metrics, we conducted an expert evaluation. All experts are authors of the paper; the majority are doctoral candidates, and all hold at least a Bachelor's degree in Swiss law. We only evaluated the laws and headnotes since they are much shorter and we could evaluate more examples in the time available. Due to limited expert time, we selected the top model from four categories: frontier (Claude-3.5-Sonnet), reasoning (o1), translation (MADLAD-400-7B) and finetuned (SLT-Qwen2.5-32B). 

The experts were asked to assign a score between 0 and 100 to each translation. For this purpose, the experts were given a source text, its “gold translation” (official translation of the Swiss authorities) as a reference and a predicted translation. The scores only reflected the completeness and accuracy of the predicted translation, with less emphasis on  readability and other stylistic attributes. To ensure consistency, the experts agreed on a point deduction system in advance and discussed certain borderline cases (annotation guidelines are in \autoref{sec:annotation-guidelines}). In total, 200 translations were assigned a score by exactly two experts. Each expert assigned scores independently, without consulting the other annotators. For the expert agreement with judge metrics (see \autoref{sec:swiltra-judge}) and for the evaluation of the best models (see \autoref{sec:best-model}), we averaged the scores of the two annotators.

\subsection{Inter-Annotator Agreement}


The average Krippendorff's $\alpha$ was 0.56 for laws and 0.41 for headnotes. Agreement was generally higher for laws than for headnotes across most language pairs and tasks. This consistent pattern is unlikely due to a calibration issue and instead suggests that the headnote task inherently allows for more subjectivity. This is likely due to headnotes being longer on average and containing greater interpretive nuance, which can lead annotators to miss or accord different importance to different details. We also hypothesize that laws inherently exhibit a more structured and clearer pattern, making them easier to translate and evaluate.

\begin{figure}[htb]
    \centering
    \includegraphics[width=\columnwidth]{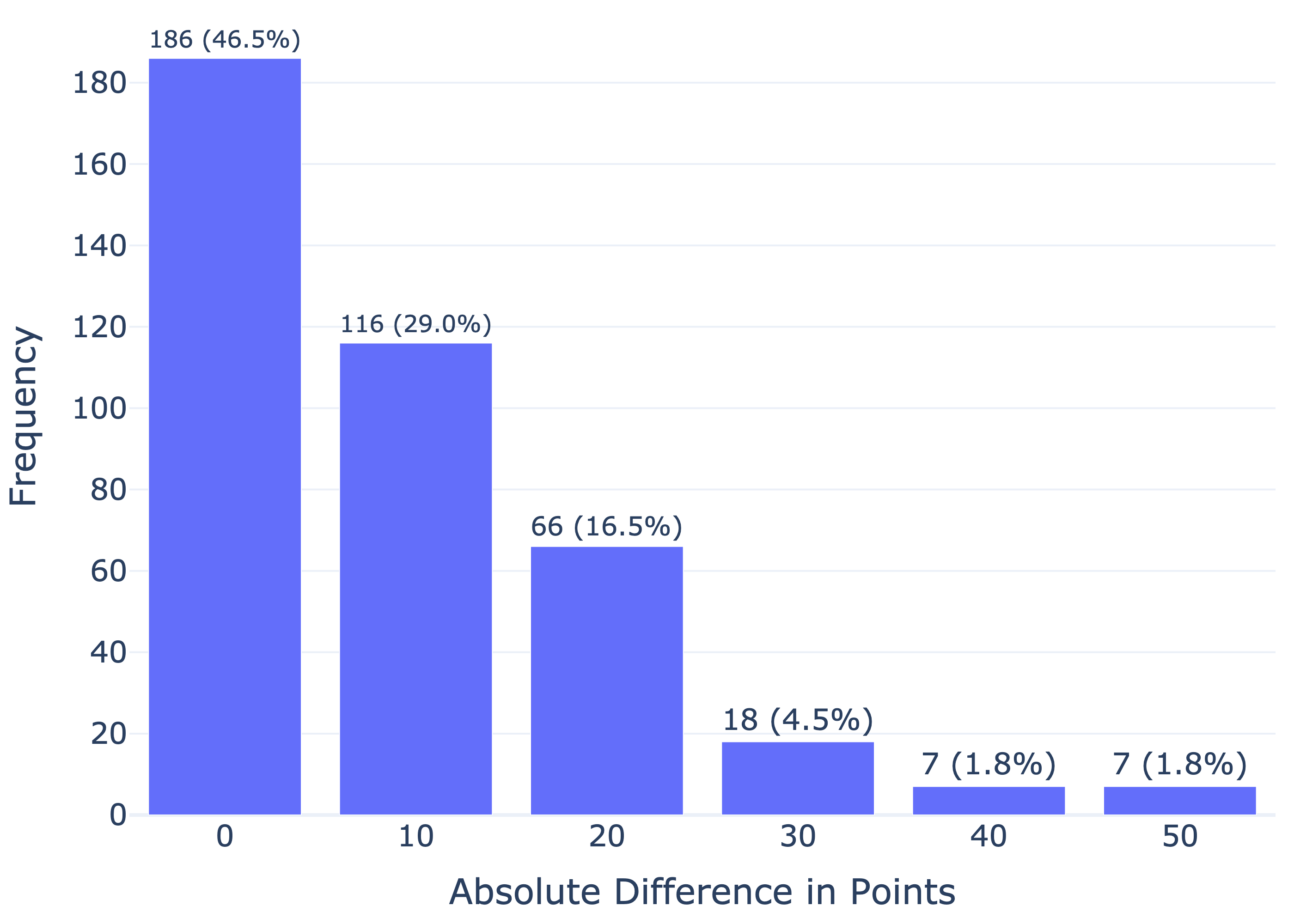}
    \caption{Absolute point difference between annotators.}
    \label{fig:annotator-differences}
\end{figure}

The moderate inter-annotator agreement suggests that, despite clear instructions, a certain degree of subjectivity was inherent in the task. In addition, we observed smaller differences between the individual language pairs, suggesting that not all annotators were perfectly aligned. However, disagreements tended to be minor and were rarely fundamental. In \autoref{fig:annotator-differences} we show the absolute point difference between the two annotators evaluating the same samples. In almost half of the cases the two annotators completely agree and in 92\% the difference is smaller than 30 points.


\subsection{Which Model is the Best?}
\label{sec:best-model}

In \autoref{tab:expert-scores} we show the expert scores together with the best metrics for the best models per category. It is evident here that XCOMET aligns best with the experts. We conclude that both for translating laws and headnotes Claude 3.5 Sonnet is the best model followed by o1 for laws and both o1 and the finetuned Qwen2.5-32B model for headnotes.

\begin{table}[!htb]
\resizebox{\columnwidth}{!}{\begin{tabular}{llrrrr}
\toprule
\textbf{Model} & \textbf{Task} & \textbf{↑ Experts} & \textbf{↑ XCOMET} & \textbf{↑ BLEURT} & \textbf{↑ GEMBA-MQM}\\
\midrule
Claude-3.5-Sonnet & Headnote & \cellcolor{teal!58}89.21 ± 2.2 & \cellcolor{teal!54}90.91 ± 1.5 & \cellcolor{teal!54}28.96 ± 3.7 & \cellcolor{teal!51}86.53 ± 1.6 \\
Claude-3.5-Sonnet & Law & \textbf{\cellcolor{teal!70}94.55 ± 1.1} & \textbf{\cellcolor{teal!70}93.30 ± 1.1} & \underline{\cellcolor{teal!65}34.16 ± 3.2} & \cellcolor{teal!57}88.86 ± 1.2 \\
MADLAD-400-7B & Headnote & \cellcolor{teal!20}71.77 ± 3.3 & \cellcolor{teal!20}85.57 ± 2.8 & \cellcolor{teal!20}12.20 ± 3.1 & \cellcolor{teal!20}76.13 ± 4.9 \\
MADLAD-400-7B & Law & \cellcolor{teal!46}83.77 ± 2.8 & \cellcolor{teal!44}89.42 ± 2.3 & \cellcolor{teal!54}28.97 ± 3.5 & \cellcolor{teal!57}88.63 ± 2.0 \\
SLT-Qwen2.5-32B & Headnote & \cellcolor{teal!48}84.86 ± 2.4 & \cellcolor{teal!39}88.62 ± 2.1 & \cellcolor{teal!58}30.78 ± 4.3 & \cellcolor{teal!20}75.89 ± 4.2 \\
SLT-Qwen2.5-32B & Law & \cellcolor{teal!50}85.74 ± 2.1 & \cellcolor{teal!35}88.03 ± 2.2 & \textbf{\cellcolor{teal!70}36.42 ± 3.9} & \cellcolor{teal!37}81.78 ± 2.5 \\
o1 & Headnote & \cellcolor{teal!47}84.29 ± 2.1 & \cellcolor{teal!45}89.58 ± 1.8 & \cellcolor{teal!29}16.77 ± 4.2 & \underline{\cellcolor{teal!68}92.34 ± 1.4} \\
o1 & Law & \underline{\cellcolor{teal!59}89.91 ± 1.5} & \underline{\cellcolor{teal!63}92.33 ± 1.5} & \cellcolor{teal!52}28.19 ± 3.2 & \textbf{\cellcolor{teal!70}92.97 ± 1.0} \\
\bottomrule
\end{tabular}}
\caption{Expert scores for best models across categories}
\label{tab:expert-scores}
\end{table}

\section{SwiLTra-Judge}
\label{sec:swiltra-judge}

Automatic evaluation of natural language generation is challenging. Lexical metrics like BLEU or METEOR correlate weakly with human judgments \cite{zhang_bertscore_2020}. Early model-based metrics such as BERTScore or BLEURT perform better, but recently, LLM-as-Judge has emerged as the dominant paradigm \cite{zheng2023judgingllmasajudgemtbenchchatbot}. Each task, however, is unique and requires its own judge setup. In this section, we ablate key aspects of the judge setup, including the judge model, prompt, and few-shot sample selection.


\subsection{Setup}
We use GPT-4o, GPT-4o-mini, Gemini-1.5-pro, and Gemini-1.5-flash in our judge model ablation. We also tested Claude Sonnet and Haiku as judges, but they failed to follow grading instructions.\footnote{They would insist on generating JSON output while we very clearly just asked for plain-text.} The o1 and o1-mini models showed very low or even negative correlations with human judgments and are thus excluded. 
We randomly selected one few-shot example from the dev sets of laws, headnotes, and press releases. To ensure judge models saw diverse translation qualities, we chose models of varying strengths (Claude 3.5 Sonnet for laws, Mixtral-8x7B-Instruct-v0.1 for headnotes, and Qwen2.5-1.5B-Instruct for press releases). We used a simple prompt ``\textit{Translate to {target-language}}'' to generate translations. Sample judgments per few-shot example were written by one lawyer author and double-checked by another. 
We tested two few-shot styles \textit{single} (all examples in one language direction: fr-de) and \textit{diverse} (law article en-it, headnote de-fr and press release fr-de). We ablated two user prompts with absolute grading (\textit{basic} and \textit{detailed}) and one with deduction grading similar to the codebook given to the human expert annotators (\textit{codebook}). 
Judge prompts are in \autoref{sec:judge-prompts}.

We measured the correlation of our judge setups with the human expert scores on the 400 human annotated samples. To get a higher confidence signal, we removed samples where the two human experts disagreed by 30 points or more (32/400 or 8\%). Find complete results in \autoref{tab:correlation-metrics-comprehensive}. Unless specified otherwise, we report Spearman correlation with human judgments with cross validation. 
Based on our expert evaluation, we answer the following research questions (RQs):

\subsection*{RQ1: Are small models judges good enough?}
\textit{A: Yes, the small models even outperform their larger counterparts.}
Over all tested configurations GPT-4o and GPT-4o-mini are tied at $0.41 \pm 0.08$ mean Spearman correlation. Gemini-1.5-flash even outperforms Gemini-1.5-pro as a judge model ($0.33 \pm 0.07$ vs $0.27 \pm 0.09$). For the best configuration GPT-4o-mini even outperforms GPT-4o ($0.48 \pm 0.1$ vs $0.45 \pm 0.07$) and the same holds for Gemini-1.5-flash vs Gemini-1.5-pro ($0.5 \pm 0.07$ vs $0.3 \pm 0.08$).


\subsection*{RQ2: Is the deduction judgment style better than the absolute style?}
\textit{A: Judges using the deduction style align more closely with human judgments.}
Across all configurations, there is little difference between the two absolute styles ($0.33 \pm 0.09$ for \textit{basic} and $0.32 \pm 0.08$ for \textit{detailed} user prompt). However, the deduction style aligns much more closely with experts ($0.42 \pm 0.08$). The top six highest correlating configurations all use the deduction style. This finding anecdotally confirms that LLM judge models reach judgments more similar to human experts when prompted in a more aligned way.

\subsection*{RQ3: Are few-shot examples in a single language pair sufficient, or is it necessary to include examples from diverse language pairs?}
\textit{A: On average, the language directions of the few shot examples do not matter, but the best configuration uses diverse language directions.}
Across all 24 investigated configurations, there is only minimal difference between the single and diverse language direction setup ($0.37 \pm 0.08$ vs $0.35 \pm 0.08$). 
However, the best configuration overall, uses diverse language directions.

\subsection*{RQ4: How does SwiLTra-Judge perform compared to other metrics?}
\begin{figure}[htb]
    \centering
    \includegraphics[width=\columnwidth]{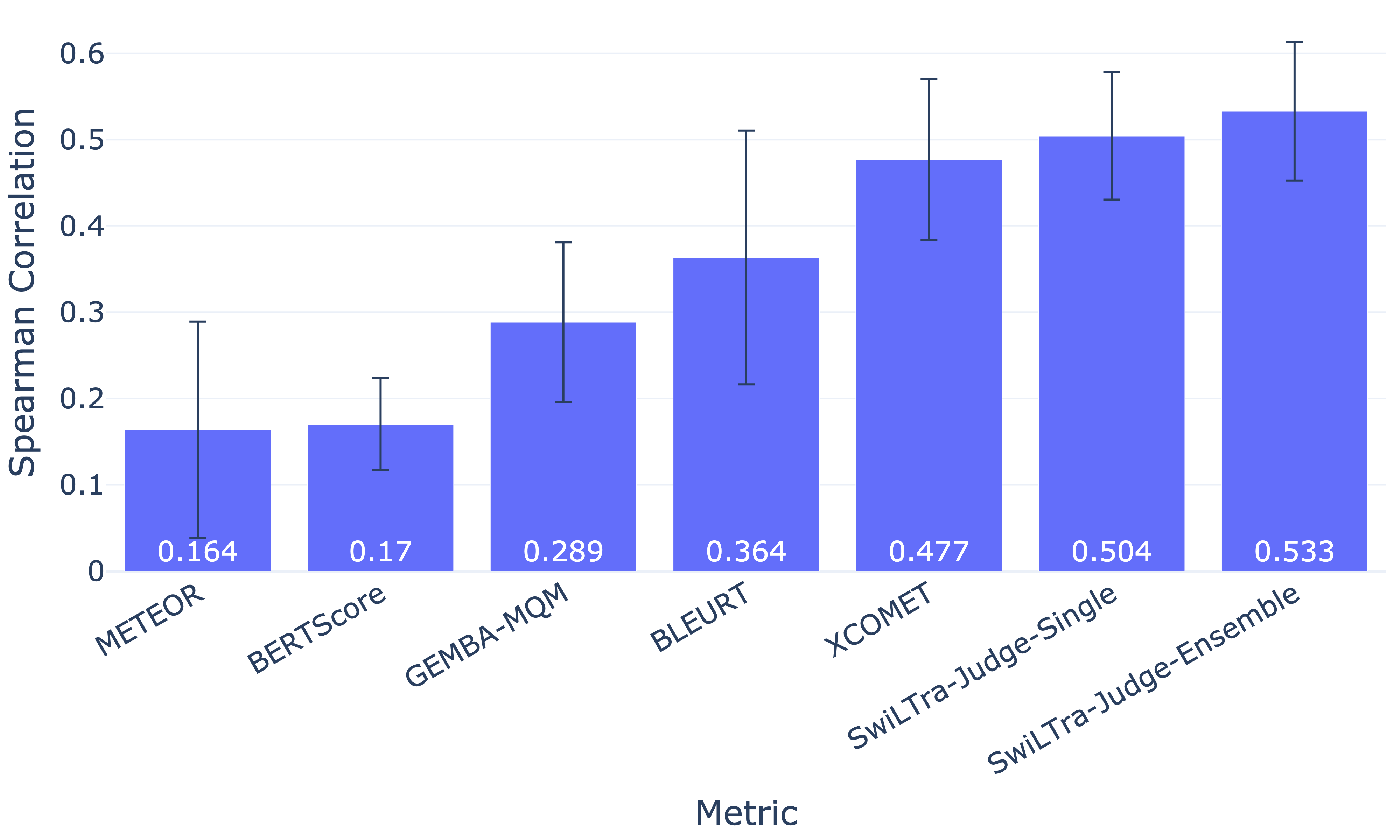}
    \caption{Spearman correlations with human expert scores.}
    \label{fig:metric-correlations}
\end{figure}

\textit{A: Our SwiLTra-Judge exhibits the highest correlation with human judgments among tested translation metrics.}
\autoref{fig:metric-correlations} shows Spearman correlation with human judgments for sample-level metrics (this excludes BLEU and ChrF). As expected, METEOR and BERTScore perform poorly, with correlations below 0.2. Surprisingly, the recent GEMBA-MQM metric both underperforms BLEURT and XCOMET. Our SwiLTra-Judge-Single (gemini-1-5-flash-codebook-diverse-deduction) is significantly better than the second-best metric XCOMET ($0.5 \pm 0.07$ vs $0.48 \pm 0.09$, $p=0.0008$). Our SwiLTra-Judge-Ensemble (a simple mean score from GPT-4o-mini and Gemini 1.5 Flash with codebook style and both single and diverse few-shot setup) is again significantly better than our SwiLTra-Judge-Single ($0.53 \pm 0.08$ vs $0.5 \pm 0.07$, $p<0.0001$). To compute both p-values, we have performed an unpaired t-test with $N = 368$ each.

%
%

\subsection{Judge Harshness}

\begin{figure}[!t]
    \centering
    \includegraphics[width=\columnwidth]{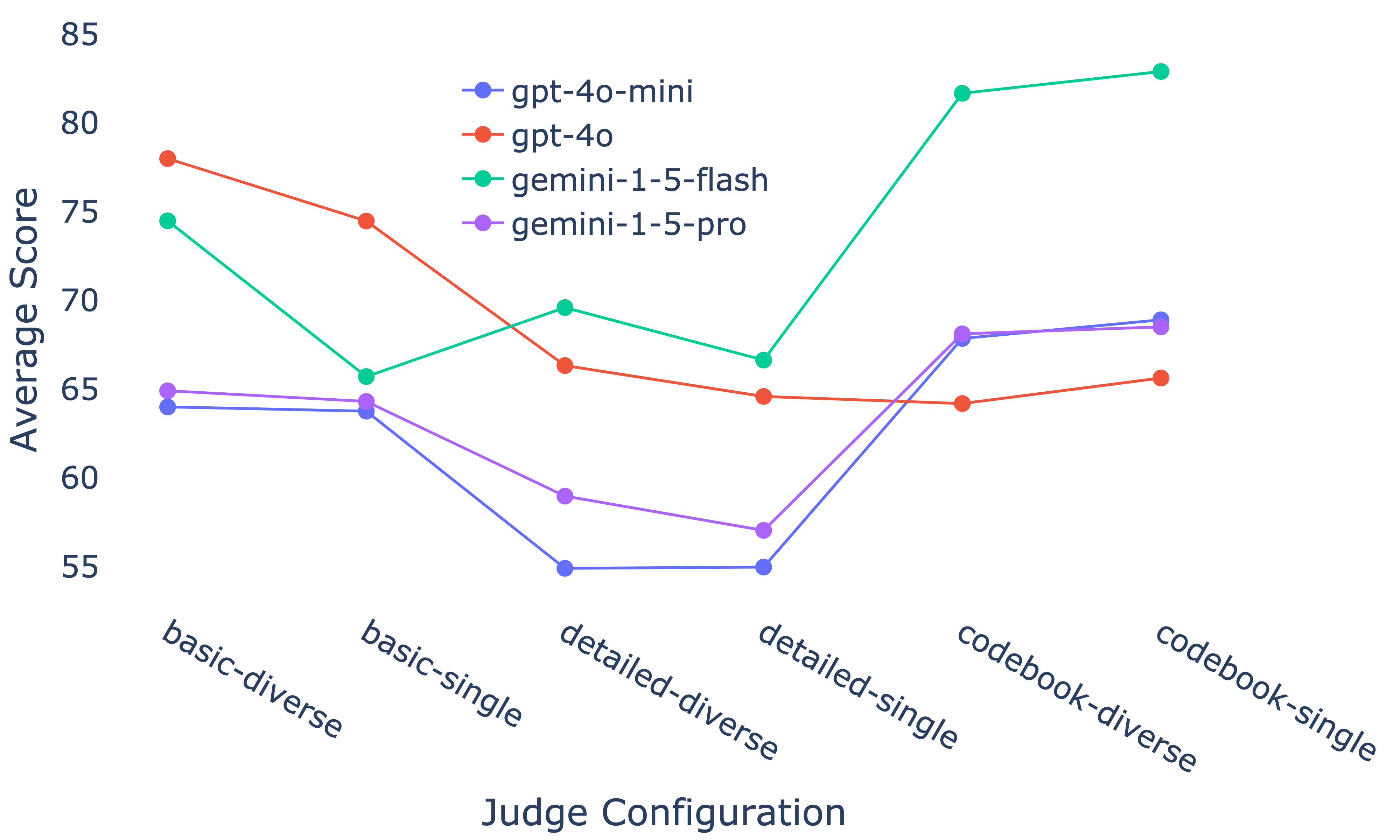}
    \caption{Judge harshness across configurations.}
    \label{fig:judge-harshness}
\end{figure}

In \autoref{fig:judge-harshness} we show the average score over the best four generator models per category across judge models, system and few shot styles. We confirm here that the language directions of the few shot examples only has a minor effect. We see that the detailed system style leads to the harshest scores across models. Interestingly, Gemini-1.5-pro and GPT-4o-mini judge very similarly in terms of harshness. All models except GPT-4o judge more leniently with the codebook system style.


\subsection{Best Model Per Task}

Now that we built a trusted metric for our translation benchmark, we ran it over the entire dataset for all models. With this, we can recommend the best model for each task.
In \autoref{fig:best-per-task} we show the top three models per task using SwiLTra-Judge as a metric. There are no large differences among top models, but the highest scores are achieved by Sonnet on headnotes, MADLAD-400-7B on laws and o1 on press releases. Sonnet ranks in the top 3 for all tasks. One reason for Sonnet’s high scores could be that the few-shot example in the judge prompt with the highest score was translated by Sonnet, possibly making the judge models prefer its style. However, human experts clearly favored Sonnet without bias from few-shot examples.
\section{Related Work}

The application of NLP to legal texts has seen significant growth in recent years. This increased attention is driven by the growing need to automate and enhance legal processes, improve access to justice, and handle the vast amounts of legal documentation produced globally.

Recent research has explored various aspects of legal text processing. Legal judgment prediction has emerged as a crucial area, with studies demonstrating success across different jurisdictions, including the US~\cite{semo_classactionprediction_2022}, Europe~\cite{vaudaux-etal-2023-pretrained} and Switzerland~\cite{DBLP:conf/acl-nllp/NiklausCS21, niklaus_empirical_2022}. Notable advances have been made in verdict prediction~\cite{DBLP:journals/ail/MedvedevaVW20}, topic classification~\cite{DBLP:conf/acl-nllp/PapaloukasCAPK21,DBLP:conf/kes/BenedettoSBTCG23, rasiah_scale_2023}, and legal QA systems~\cite{DBLP:conf/aaai/ZhongXTZ0S20}. LegalBench~\cite{guha_legalbench_2023} LexGLUE~\cite{DBLP:conf/acl/ChalkidisJHBAKA22} and LEXTREME~\cite{niklaus_lextreme_2023} are established as comprehensive benchmark suites comprising multiple legal NLP tasks, including text classification, named entity recognition, and legal entailment across various jurisdictions and legal areas.

The translation of legal texts has significant societal impact and is increasingly important for training translators and practical applications, especially as machine translation gains prominence~\cite{killman2024machine}. However, legal translation poses challenges due to domain-specific terminology, reliability in legal formulae, and non-compliance with legal conventions~\cite{killman2023machine, giampieri2023machine}. While some U.S. courts have considered NMT, it remains far from replacing human translators~\cite{vieira2021understanding}.
Robust NMT systems are essential for judicial and governmental services, with recent advancements leveraging pretrained LLMs and fine-tuning techniques~\cite{DBLP:conf/naacl/ZhuLDXHKCL24}. Prior research has focused on legal NMT for languages like Chinese, and Arabic~\cite{ding2024comparative, elfqih2023evaluation}. However, a significant research gap remains in translating legal texts between Switzerland’s national languages, which our work aims to address.
\section{Conclusions}
\label{sec:conclusions}

In this work, we introduced \textsc{SwiLTra-Bench}, a high-quality multilingual legal translation benchmark, and evaluated mainstream LLM-based NMT systems under both zero-shot and fine-tuned settings. Our analysis, validated by human expert annotations, showed that frontier models outperform all others, while translation-specific systems like MADLAD-400 excel on laws but struggle with headnotes. Fine-tuning open LLMs significantly improves their performance, though they still lag behind zero-shot frontier models, and translation quality remains consistent across Swiss languages. Finally, our \textsc{SwiLTra-Judge} model, optimized for legal translation evaluation, achieves the highest alignment with human expert judgments, providing a valuable automated metric for future research.
\section*{Limitations \& Future Work}
\label{sec:limitations-future-work}

Our fine-tuned models are much stronger than the initial instruction-tuned open models they are based on, but they still under-perform large closed models. Future work could investigate techniques such as model merging \cite{yang2024modelmergingllmsmllms} to further improve and bring them closer to the frontier models.
While we evaluated a large variety of models, we could not evaluate them all. Future work could investigate other promising models such as Grok.\footnote{\href{https://x.ai/blog/grok-2}{https://x.ai/blog/grok-2}}
We took great care to validate our results with human expert studies. However, our resources were limited and we could not investigate certain languages (e.g., Romansh) and our sample sizes were still rather small. Future work could perform a more broad and in-depth human evaluation.


\section*{Ethics Statement}

Our benchmark contains no personal, sensitive, or private information; it consists solely of publicly available data.


\section*{Acknowledgments}
We thank the anonymous reviewers for their constructive feedback. Luka Nenadic’s research is funded by SNSF grant No. 10002634.

\bibliography{references,custom}

\begin{thebibliography}{59}
\providecommand{\natexlab}[1]{#1}

\bibitem[{Abdin et~al.(2024)Abdin, Aneja, Awadalla, Awadallah, Awan, and \textit{et al.}}]{phi3shortened}
Marah Abdin, Jyoti Aneja, Hany Awadalla, Ahmed Awadallah, Ammar~Ahmad Awan, and \textit{et al.} 2024.
\newblock \href {https://arxiv.org/abs/2404.14219} {Phi-3 technical report: A highly capable language model locally on your phone}.
\newblock \emph{Preprint}, arXiv:2404.14219.

\bibitem[{Alves et~al.(2023)Alves, Guerreiro, Alves, Pombal, Rei, de~Souza, Colombo, and Martins}]{alves2023steering}
Duarte Alves, Nuno Guerreiro, Jo{\~a}o Alves, Jos{\'e} Pombal, Ricardo Rei, Jos{\'e} de~Souza, Pierre Colombo, and Andr{\'e}~FT Martins. 2023.
\newblock Steering large language models for machine translation with finetuning and in-context learning.
\newblock In \emph{Findings of the Association for Computational Linguistics: EMNLP 2023}, pages 11127--11148.

\bibitem[{Alves et~al.(2024)Alves, Pombal, Guerreiro, Martins, Alves, Farajian, Peters, Rei, Fernandes, Agrawal, Colombo, de~Souza, and Martins}]{tower}
Duarte~M. Alves, José Pombal, Nuno~M. Guerreiro, Pedro~H. Martins, João Alves, Amin Farajian, Ben Peters, Ricardo Rei, Patrick Fernandes, Sweta Agrawal, Pierre Colombo, José G.~C. de~Souza, and André F.~T. Martins. 2024.
\newblock \href {https://arxiv.org/abs/2402.17733} {Tower: An open multilingual large language model for translation-related tasks}.
\newblock \emph{Preprint}, arXiv:2402.17733.

\bibitem[{Banerjee and Lavie(2005)}]{banerjee_meteor_2005}
Satanjeev Banerjee and Alon Lavie. 2005.
\newblock \href {https://aclanthology.org/W05-0909} {{METEOR}: {An} {Automatic} {Metric} for {MT} {Evaluation} with {Improved} {Correlation} with {Human} {Judgments}}.
\newblock In \emph{Proceedings of the {ACL} {Workshop} on {Intrinsic} and {Extrinsic} {Evaluation} {Measures} for {Machine} {Translation} and/or {Summarization}}, pages 65--72, Ann Arbor, Michigan. Association for Computational Linguistics.

\bibitem[{Benedetto et~al.(2023)Benedetto, Sportelli, Bertoldo, Tarasconi, Cagliero, and Giacalone}]{DBLP:conf/kes/BenedettoSBTCG23}
Irene Benedetto, Gianpiero Sportelli, Sara Bertoldo, Francesco Tarasconi, Luca Cagliero, and Giuseppe Giacalone. 2023.
\newblock \href {https://doi.org/10.1016/J.PROCS.2023.10.215} {On the use of pretrained language models for legal {I}talian document classification}.
\newblock In \emph{Knowledge-Based and Intelligent Information {\&} Engineering Systems: Proceedings of the 27th International Conference KES-2023, Athens, Greece, 6-8 September 2023}, volume 225 of \emph{Procedia Computer Science}, pages 2244--2253. Elsevier.

\bibitem[{Brown et~al.(2020)Brown, Mann, Ryder, Subbiah, Kaplan, Dhariwal, Neelakantan, Shyam, Sastry, Askell, Agarwal, Herbert{-}Voss, Krueger, Henighan, Child, Ramesh, Ziegler, Wu, Winter, Hesse, Chen, Sigler, Litwin, Gray, Chess, Clark, Berner, McCandlish, Radford, Sutskever, and Amodei}]{brown2020language}
Tom~B. Brown, Benjamin Mann, Nick Ryder, Melanie Subbiah, Jared Kaplan, Prafulla Dhariwal, Arvind Neelakantan, Pranav Shyam, Girish Sastry, Amanda Askell, Sandhini Agarwal, Ariel Herbert{-}Voss, Gretchen Krueger, Tom Henighan, Rewon Child, Aditya Ramesh, Daniel~M. Ziegler, Jeffrey Wu, Clemens Winter, Christopher Hesse, Mark Chen, Eric Sigler, Mateusz Litwin, Scott Gray, Benjamin Chess, Jack Clark, Christopher Berner, Sam McCandlish, Alec Radford, Ilya Sutskever, and Dario Amodei. 2020.
\newblock \href {https://proceedings.neurips.cc/paper/2020/hash/1457c0d6bfcb4967418bfb8ac142f64a-Abstract.html} {Language models are few-shot learners}.
\newblock In \emph{Advances in Neural Information Processing Systems 33: Annual Conference on Neural Information Processing Systems 2020, NeurIPS 2020, December 6-12, 2020, virtual}.

\bibitem[{Canavese and Cadwell(2024)}]{canavese2024translators}
Paolo Canavese and Patrick Cadwell. 2024.
\newblock Translators’ perspectives on machine translation uses and impacts in the {Swiss Confederation}: Navigating technological change in an institutional setting.
\newblock In \emph{Proceedings of the 25th Annual Conference of the European Association for Machine Translation (Volume 1)}, pages 347--359.

\bibitem[{Chalkidis et~al.(2022)Chalkidis, Jana, Hartung, II, Androutsopoulos, Katz, and Aletras}]{DBLP:conf/acl/ChalkidisJHBAKA22}
Ilias Chalkidis, Abhik Jana, Dirk Hartung, Michael J.~Bommarito II, Ion Androutsopoulos, Daniel~Martin Katz, and Nikolaos Aletras. 2022.
\newblock \href {https://doi.org/10.18653/V1/2022.ACL-LONG.297} {{LexGLUE: A Benchmark Dataset for Legal Language Understanding in English}}.
\newblock In \emph{Proceedings of the 60th Annual Meeting of the Association for Computational Linguistics (Volume 1: Long Papers), {ACL} 2022, Dublin, Ireland, May 22-27, 2022}, pages 4310--4330. Association for Computational Linguistics.

\bibitem[{Communication et~al.(2023)Communication, Barrault, Chung, Meglioli, Dale, Dong, Duquenne, Elsahar, Gong, Heffernan, Hoffman, Klaiber, Li, Licht, Maillard, Rakotoarison, Sadagopan, Wenzek, Ye, Akula, Chen, Hachem, Ellis, Gonzalez, Haaheim, Hansanti, Howes, Huang, Hwang, Inaguma, Jain, Kalbassi, Kallet, Kulikov, Lam, Li, Ma, Mavlyutov, Peloquin, Ramadan, Ramakrishnan, Sun, Tran, Tran, Tufanov, Vogeti, Wood, Yang, Yu, Andrews, Balioglu, Costa-jussà, Celebi, Elbayad, Gao, Guzmán, Kao, Lee, Mourachko, Pino, Popuri, Ropers, Saleem, Schwenk, Tomasello, Wang, Wang, and Wang}]{seamlessm4t}
Seamless Communication, Loïc Barrault, Yu-An Chung, Mariano~Cora Meglioli, David Dale, Ning Dong, Paul-Ambroise Duquenne, Hady Elsahar, Hongyu Gong, Kevin Heffernan, John Hoffman, Christopher Klaiber, Pengwei Li, Daniel Licht, Jean Maillard, Alice Rakotoarison, Kaushik~Ram Sadagopan, Guillaume Wenzek, Ethan Ye, Bapi Akula, Peng-Jen Chen, Naji~El Hachem, Brian Ellis, Gabriel~Mejia Gonzalez, Justin Haaheim, Prangthip Hansanti, Russ Howes, Bernie Huang, Min-Jae Hwang, Hirofumi Inaguma, Somya Jain, Elahe Kalbassi, Amanda Kallet, Ilia Kulikov, Janice Lam, Daniel Li, Xutai Ma, Ruslan Mavlyutov, Benjamin Peloquin, Mohamed Ramadan, Abinesh Ramakrishnan, Anna Sun, Kevin Tran, Tuan Tran, Igor Tufanov, Vish Vogeti, Carleigh Wood, Yilin Yang, Bokai Yu, Pierre Andrews, Can Balioglu, Marta~R. Costa-jussà, Onur Celebi, Maha Elbayad, Cynthia Gao, Francisco Guzmán, Justine Kao, Ann Lee, Alexandre Mourachko, Juan Pino, Sravya Popuri, Christophe Ropers, Safiyyah Saleem, Holger Schwenk, Paden Tomasello, Changhan Wang, Jeff
  Wang, and Skyler Wang. 2023.
\newblock \href {https://arxiv.org/abs/2308.11596} {Seamlessm4t: Massively multilingual \& multimodal machine translation}.
\newblock \emph{Preprint}, arXiv:2308.11596.

\bibitem[{Dai and Le(2015)}]{dai2015semi}
Andrew~M. Dai and Quoc~V. Le. 2015.
\newblock \href {https://proceedings.neurips.cc/paper/2015/hash/7137debd45ae4d0ab9aa953017286b20-Abstract.html} {Semi-supervised sequence learning}.
\newblock In \emph{Advances in Neural Information Processing Systems 28: Annual Conference on Neural Information Processing Systems 2015, December 7-12, 2015, Montreal, Quebec, Canada}, pages 3079--3087.

\bibitem[{Dettmers et~al.(2022)Dettmers, Lewis, Shleifer, and Zettlemoyer}]{dettmers20228bitoptimizersblockwisequantization}
Tim Dettmers, Mike Lewis, Sam Shleifer, and Luke Zettlemoyer. 2022.
\newblock \href {https://openreview.net/forum?id=shpkpVXzo3h} {8-bit optimizers via block-wise quantization}.
\newblock In \emph{The Tenth International Conference on Learning Representations, {ICLR} 2022, Virtual Event, April 25-29, 2022}. OpenReview.net.

\bibitem[{Ding(2024)}]{ding2024comparative}
Lijie Ding. 2024.
\newblock {A Comparative Study on the Quality of English-Chinese Translation of Legal Texts Between ChatGPT and Neural Machine Translation Systems}.
\newblock \emph{Theory and Practice in Language Studies}, 14(9):2823--2833.

\bibitem[{ElFqih and Monti(2023)}]{elfqih2023evaluation}
Khadija~Ait ElFqih and Johanna Monti. 2023.
\newblock {On the Evaluation of Terminology Translation Errors in NMT and PB-SMT In the Legal Domain: A Study on the Translation of Arabic Legal Documents into English and French}.
\newblock In \emph{Proceedings of the Workshop on Computational Terminology in NLP and Translation Studies (ConTeNTS) Incorporating the 16th Workshop on Building and Using Comparable Corpora (BUCC)}, pages 26--35.

\bibitem[{Giampieri(2023)}]{giampieri2023machine}
Patrizia Giampieri. 2023.
\newblock Is machine translation reliable in the legal field? a corpus-based critical comparative analysis for teaching {ESP} at tertiary level.
\newblock \emph{Esp Today-Journal of English for Specific Purposes at Tertiary Level}, 11(1):119--137.

\bibitem[{Grattafiori et~al.(2024)Grattafiori, Dubey, Jauhri, Pandey, Kadian, and \textit{et al.}}]{llama3shortened}
Aaron Grattafiori, Abhimanyu Dubey, Abhinav Jauhri, Abhinav Pandey, Abhishek Kadian, and \textit{et al.} 2024.
\newblock The {L}lama 3 herd of models.
\newblock \emph{arXiv preprint arXiv:2404.14219}.

\bibitem[{Guerreiro et~al.(2024)Guerreiro, Rei, Stigt, Coheur, Colombo, and Martins}]{guerreiro-xcomet}
Nuno~M. Guerreiro, Ricardo Rei, Daan~van Stigt, Luisa Coheur, Pierre Colombo, and André F.~T. Martins. 2024.
\newblock \href {https://doi.org/10.1162/tacl_a_00683} {{xCOMET}: Transparent machine translation evaluation through fine-grained error detection}.
\newblock \emph{Transactions of the Association for Computational Linguistics}, 12:979--995.

\bibitem[{Guha et~al.(2023)Guha, Nyarko, Ho, Ré, Chilton, Narayana, Chohlas-Wood, Peters, Waldon, Rockmore, Zambrano, Talisman, Hoque, Surani, Fagan, Sarfaty, Dickinson, Porat, Hegland, Wu, Nudell, Niklaus, Nay, Choi, Tobia, Hagan, Ma, Livermore, Rasumov-Rahe, Holzenberger, Kolt, Henderson, Rehaag, Goel, Gao, Williams, Gandhi, Zur, Iyer, and Li}]{guha_legalbench_2023}
Neel Guha, Julian Nyarko, Daniel~E. Ho, Christopher Ré, Adam Chilton, Aditya Narayana, Alex Chohlas-Wood, Austin Peters, Brandon Waldon, Daniel~N. Rockmore, Diego Zambrano, Dmitry Talisman, Enam Hoque, Faiz Surani, Frank Fagan, Galit Sarfaty, Gregory~M. Dickinson, Haggai Porat, Jason Hegland, Jessica Wu, Joe Nudell, Joel Niklaus, John Nay, Jonathan~H. Choi, Kevin Tobia, Margaret Hagan, Megan Ma, Michael Livermore, Nikon Rasumov-Rahe, Nils Holzenberger, Noam Kolt, Peter Henderson, Sean Rehaag, Sharad Goel, Shang Gao, Spencer Williams, Sunny Gandhi, Tom Zur, Varun Iyer, and Zehua Li. 2023.
\newblock \href {https://doi.org/10.48550/arXiv.2308.11462} {{LegalBench}: {A} {Collaboratively} {Built} {Benchmark} for {Measuring} {Legal} {Reasoning} in {Large} {Language} {Models}}.
\newblock \emph{arXiv preprint}.
\newblock ArXiv:2308.11462 [cs].

\bibitem[{Han et~al.(2024)Han, Gladkoff, Erofeev, Sorokina, Galiano, and Nenadic}]{han2024neural}
Lifeng Han, Serge Gladkoff, Gleb Erofeev, Irina Sorokina, Betty Galiano, and Goran Nenadic. 2024.
\newblock Neural machine translation of clinical text: an empirical investigation into multilingual pre-trained language models and transfer-learning.
\newblock \emph{Frontiers in Digital Health}, 6:1211564.

\bibitem[{Hu et~al.(2021)Hu, Shen, Wallis, Allen-Zhu, Li, Wang, Wang, and Chen}]{hu_lora_2021}
Edward~J. Hu, Yelong Shen, Phillip Wallis, Zeyuan Allen-Zhu, Yuanzhi Li, Shean Wang, Lu~Wang, and Weizhu Chen. 2021.
\newblock \href {http://arxiv.org/abs/2106.09685} {{LoRA}: {Low}-{Rank} {Adaptation} of {Large} {Language} {Models}}.
\newblock \emph{arXiv preprint}.
\newblock ArXiv:2106.09685 [cs].

\bibitem[{Kalajdzievski(2023)}]{kalajdzievski2023rankstabilizationscalingfactor}
Damjan Kalajdzievski. 2023.
\newblock \href {https://doi.org/10.48550/ARXIV.2312.03732} {A rank stabilization scaling factor for fine-tuning with {LoRA}}.
\newblock \emph{CoRR}, abs/2312.03732.

\bibitem[{Katz et~al.(2023)Katz, Hartung, Gerlach, Jana, and Bommarito~II}]{katz2023natural}
Daniel~Martin Katz, Dirk Hartung, Lauritz Gerlach, Abhik Jana, and Michael~J Bommarito~II. 2023.
\newblock Natural language processing in the legal domain.
\newblock \emph{arXiv preprint arXiv:2302.12039}.

\bibitem[{Killman(2023)}]{killman2023machine}
Jeffrey Killman. 2023.
\newblock Machine translation and legal terminology: Data-driven approaches to contextual accuracy.
\newblock \emph{Handbook of Terminology}, pages 485--510.

\bibitem[{Killman(2024)}]{killman2024machine}
Jeffrey Killman. 2024.
\newblock Machine translation literacy in the legal translation context: a {SWOT} analysis perspective.
\newblock \emph{The Interpreter and Translator Trainer}, 18(2):271--289.

\bibitem[{Kocmi and Federmann(2023)}]{kocmi-federmann-2023-gemba}
Tom Kocmi and Christian Federmann. 2023.
\newblock \href {https://doi.org/10.18653/v1/2023.wmt-1.64} {{GEMBA}-{MQM}: Detecting translation quality error spans with {GPT}-4}.
\newblock In \emph{Proceedings of the Eighth Conference on Machine Translation}, pages 768--775, Singapore. Association for Computational Linguistics.

\bibitem[{Koehn(2005)}]{koehn2005europarl}
Philipp Koehn. 2005.
\newblock Europarl: A parallel corpus for statistical machine translation.
\newblock In \emph{Proceedings of {Machine Translation Summit X}: papers}, pages 79--86.

\bibitem[{Kudugunta et~al.(2024)Kudugunta, Caswell, Zhang, Garcia, Xin, Kusupati, Stella, Bapna, and Firat}]{madlad400}
Sneha Kudugunta, Isaac Caswell, Biao Zhang, Xavier Garcia, Derrick Xin, Aditya Kusupati, Romi Stella, Ankur Bapna, and Orhan Firat. 2024.
\newblock Madlad-400: a multilingual and document-level large audited dataset.
\newblock In \emph{Proceedings of the 37th International Conference on Neural Information Processing Systems}, NIPS '23, Red Hook, NY, USA. Curran Associates Inc.

\bibitem[{Loshchilov and Hutter(2019)}]{loshchilov2019decoupledweightdecayregularization}
Ilya Loshchilov and Frank Hutter. 2019.
\newblock \href {https://openreview.net/forum?id=Bkg6RiCqY7} {Decoupled weight decay regularization}.
\newblock In \emph{7th International Conference on Learning Representations, {ICLR} 2019, New Orleans, LA, USA, May 6-9, 2019}. OpenReview.net.

\bibitem[{Mart{\'\i}nez-Dom{\'\i}nguez et~al.(2020)Mart{\'\i}nez-Dom{\'\i}nguez, Rikters, Vasi{\c{l}}evskis, Pinnis, and Reichenberg}]{martinez2020customized}
Rub{\'e}n Mart{\'\i}nez-Dom{\'\i}nguez, Mat{\=\i}ss Rikters, Art{\=u}rs Vasi{\c{l}}evskis, M{\=a}rcis Pinnis, and Paula Reichenberg. 2020.
\newblock Customized neural machine translation systems for the {S}wiss legal domain.
\newblock In \emph{Proceedings of the 14th Conference of the Association for Machine Translation in the Americas (Volume 2: User Track)}, pages 217--223.

\bibitem[{Medvedeva et~al.(2020)Medvedeva, Vols, and Wieling}]{DBLP:journals/ail/MedvedevaVW20}
Masha Medvedeva, Michel Vols, and Martijn Wieling. 2020.
\newblock \href {https://doi.org/10.1007/S10506-019-09255-Y} {{Using machine learning to predict decisions of the European Court of Human Rights}}.
\newblock \emph{Artif. Intell. Law}, 28(2):237--266.

\bibitem[{Moniz and Escart{\'\i}n(2023)}]{moniz2023towards}
Helena Moniz and Carla~Parra Escart{\'\i}n. 2023.
\newblock Towards responsible machine translation.
\newblock \emph{Ethical and Legal Considerations in Machine Translation. Cham: Springer}.

\bibitem[{Moslem et~al.(2023)Moslem, Haque, Kelleher, and Way}]{moslem2023adaptive}
Yasmin Moslem, Rejwanul Haque, John Kelleher, and Andy Way. 2023.
\newblock Adaptive machine translation with large language models.
\newblock In \emph{Proceedings of the 24th Annual Conference of the European Association for Machine Translation}, pages 227--237.

\bibitem[{Niklaus et~al.(2021)Niklaus, Chalkidis, and St{\"{u}}rmer}]{DBLP:conf/acl-nllp/NiklausCS21}
Joel Niklaus, Ilias Chalkidis, and Matthias St{\"{u}}rmer. 2021.
\newblock \href {https://aclanthology.org/2021.nllp-1.3} {{Swiss-Judgment-Prediction}: {A} multilingual legal judgment prediction benchmark}.
\newblock In \emph{Proceedings of the Natural Legal Language Processing Workshop 2021, NLLP@EMNLP 2021, Punta Cana, Dominican Republic, November 10, 2021}, pages 19--35. Association for Computational Linguistics.

\bibitem[{Niklaus et~al.(2023)Niklaus, Matoshi, Rani, Galassi, Stürmer, and Chalkidis}]{niklaus_lextreme_2023}
Joel Niklaus, Veton Matoshi, Pooja Rani, Andrea Galassi, Matthias Stürmer, and Ilias Chalkidis. 2023.
\newblock \href {https://doi.org/10.18653/v1/2023.findings-emnlp.200} {{LEXTREME}: {A} {Multi}-{Lingual} and {Multi}-{Task} {Benchmark} for the {Legal} {Domain}}.
\newblock In \emph{Findings of the {Association} for {Computational} {Linguistics}: {EMNLP} 2023}, pages 3016--3054, Singapore. Association for Computational Linguistics.

\bibitem[{Niklaus et~al.(2022)Niklaus, Stürmer, and Chalkidis}]{niklaus_empirical_2022}
Joel Niklaus, Matthias Stürmer, and Ilias Chalkidis. 2022.
\newblock \href {https://aclanthology.org/2022.aacl-main.3} {An {Empirical} {Study} on {Cross}-{X} {Transfer} for {Legal} {Judgment} {Prediction}}.
\newblock In \emph{Proceedings of the 2nd {Conference} of the {Asia}-{Pacific} {Chapter} of the {Association} for {Computational} {Linguistics} and the 12th {International} {Joint} {Conference} on {Natural} {Language} {Processing} ({Volume} 1: {Long} {Papers})}, pages 32--46, Online only. Association for Computational Linguistics.

\bibitem[{Niklaus et~al.(2024)Niklaus, Zheng, McCarthy, Hahn, Rosen, Henderson, Ho, Honke, Liang, and Manning}]{niklaus_flawn-t5_2024}
Joel Niklaus, Lucia Zheng, Arya~D. McCarthy, Christopher Hahn, Brian~M. Rosen, Peter Henderson, Daniel~E. Ho, Garrett Honke, Percy Liang, and Christopher Manning. 2024.
\newblock \href {http://arxiv.org/abs/2404.02127} {{FLawN}-{T5}: {An} {Empirical} {Examination} of {Effective} {Instruction}-{Tuning} {Data} {Mixtures} for {Legal} {Reasoning}}.
\newblock \emph{arXiv preprint}.
\newblock ArXiv:2404.02127 [cs].

\bibitem[{Oliver et~al.(2024)Oliver, Alvarez-Vidal, Stemle, and Chiocchetti}]{oliver2024training}
Antoni Oliver, Sergi Alvarez-Vidal, Egon Stemle, and Elena Chiocchetti. 2024.
\newblock {Training an NMT system for legal texts of a low-resource language variety South Tyrolean German-Italian}.
\newblock In \emph{Proceedings of the 25th Annual Conference of the European Association for Machine Translation (Volume 1)}, pages 573--579.

\bibitem[{Ou et~al.(2023)Ou, Ma, Kan, and Wang}]{ou2023songs}
Longshen Ou, Xichu Ma, Min-Yen Kan, and Ye~Wang. 2023.
\newblock Songs across borders: Singable and controllable neural lyric translation.
\newblock In \emph{Proceedings of the 61st Annual Meeting of the Association for Computational Linguistics (Volume 1: Long Papers)}, pages 447--467.

\bibitem[{Panickssery et~al.(2024)Panickssery, Bowman, and Feng}]{panickssery2024llmevaluatorsrecognizefavor}
Arjun Panickssery, Samuel~R. Bowman, and Shi Feng. 2024.
\newblock \href {https://arxiv.org/abs/2404.13076} {{LLM} evaluators recognize and favor their own generations}.
\newblock \emph{Preprint}, arXiv:2404.13076.

\bibitem[{Papaloukas et~al.(2021)Papaloukas, Chalkidis, Athinaios, Pantazi, and Koubarakis}]{DBLP:conf/acl-nllp/PapaloukasCAPK21}
Christos Papaloukas, Ilias Chalkidis, Konstantinos Athinaios, Despina{-}Athanasia Pantazi, and Manolis Koubarakis. 2021.
\newblock \href {https://aclanthology.org/2021.nllp-1.6} {Multi-granular legal topic classification on {G}reek legislation}.
\newblock In \emph{Proceedings of the Natural Legal Language Processing Workshop 2021, NLLP@EMNLP 2021, Punta Cana, Dominican Republic, November 10, 2021}, pages 63--75. Association for Computational Linguistics.

\bibitem[{Papineni et~al.(2002)Papineni, Roukos, Ward, and Zhu}]{papineni_bleu_2002}
Kishore Papineni, Salim Roukos, Todd Ward, and Wei-Jing Zhu. 2002.
\newblock \href {https://doi.org/10.3115/1073083.1073135} {{BLEU}: {A} {Method} for {Automatic} {Evaluation} of {Machine} {Translation}}.
\newblock In \emph{Proceedings of the 40th {Annual} {Meeting} on {Association} for {Computational} {Linguistics}}, {ACL} '02, pages 311--318, USA. Association for Computational Linguistics.
\newblock Event-place: Philadelphia, Pennsylvania.

\bibitem[{Popovi{\'c}(2015)}]{popovic-2015-chrf}
Maja Popovi{\'c}. 2015.
\newblock \href {https://doi.org/10.18653/v1/W15-3049} {chr{F}: character n-gram {F}-score for automatic {MT} evaluation}.
\newblock In \emph{Proceedings of the Tenth Workshop on Statistical Machine Translation}, pages 392--395, Lisbon, Portugal. Association for Computational Linguistics.

\bibitem[{Rasiah et~al.(2023)Rasiah, Stern, Matoshi, Stürmer, Chalkidis, Ho, and Niklaus}]{rasiah_scale_2023}
Vishvaksenan Rasiah, Ronja Stern, Veton Matoshi, Matthias Stürmer, Ilias Chalkidis, Daniel~E. Ho, and Joel Niklaus. 2023.
\newblock \href {https://doi.org/10.48550/arXiv.2306.09237} {{SCALE}: {Scaling} up the {Complexity} for {Advanced} {Language} {Model} {Evaluation}}.
\newblock \emph{arXiv preprint}.
\newblock ArXiv:2306.09237 [cs].

\bibitem[{Sellam et~al.(2020)Sellam, Das, and Parikh}]{sellam-etal-2020-bleurt}
Thibault Sellam, Dipanjan Das, and Ankur Parikh. 2020.
\newblock \href {https://doi.org/10.18653/v1/2020.acl-main.704} {{BLEURT}: Learning robust metrics for text generation}.
\newblock In \emph{Proceedings of the 58th Annual Meeting of the Association for Computational Linguistics}, pages 7881--7892, Online. Association for Computational Linguistics.

\bibitem[{Semo et~al.(2022)Semo, Bernsohn, Hagag, Hayat, and Niklaus}]{semo_classactionprediction_2022}
Gil Semo, Dor Bernsohn, Ben Hagag, Gila Hayat, and Joel Niklaus. 2022.
\newblock \href {https://aclanthology.org/2022.nllp-1.3} {{ClassActionPrediction}: {A} {Challenging} {Benchmark} for {Legal} {Judgment} {Prediction} of {Class} {Action} {Cases} in the {US}}.
\newblock In \emph{Proceedings of the {Natural} {Legal} {Language} {Processing} {Workshop} 2022}, pages 31--46, Abu Dhabi, United Arab Emirates (Hybrid). Association for Computational Linguistics.

\bibitem[{Team et~al.(2024)Team, Riviere, Pathak, Sessa, Hardin, and \textit{et al.}}]{gemma2shortened}
Gemma Team, Morgane Riviere, Shreya Pathak, Pier~Giuseppe Sessa, Cassidy Hardin, and \textit{et al.} 2024.
\newblock \href {https://arxiv.org/abs/2408.00118} {Gemma 2: Improving open language models at a practical size}.
\newblock \emph{Preprint}, arXiv:2408.00118.

\bibitem[{Team(2024)}]{qwen2.5}
Qwen Team. 2024.
\newblock \href {https://qwenlm. github. io/blog/qwen2} {Qwen2. 5: A party of foundation models}.
\newblock \emph{Qwen (Sept. 2024).}, 5.

\bibitem[{Touvron et~al.(2023)Touvron, Lavril, Izacard, Martinet, Lachaux, Lacroix, Rozi{\`e}re, Goyal, Hambro, Azhar et~al.}]{touvron2023llama}
Hugo Touvron, Thibaut Lavril, Gautier Izacard, Xavier Martinet, Marie-Anne Lachaux, Timoth{\'e}e Lacroix, Baptiste Rozi{\`e}re, Naman Goyal, Eric Hambro, Faisal Azhar, et~al. 2023.
\newblock Llama: Open and efficient foundation language models.
\newblock \emph{arXiv preprint arXiv:2302.13971}.

\bibitem[{Vaswani et~al.(2017)Vaswani, Shazeer, Parmar, Uszkoreit, Jones, Gomez, Kaiser, and Polosukhin}]{vaswani2017attention}
Ashish Vaswani, Noam Shazeer, Niki Parmar, Jakob Uszkoreit, Llion Jones, Aidan~N. Gomez, Lukasz Kaiser, and Illia Polosukhin. 2017.
\newblock \href {https://proceedings.neurips.cc/paper/2017/hash/3f5ee243547dee91fbd053c1c4a845aa-Abstract.html} {Attention is all you need}.
\newblock In \emph{Advances in Neural Information Processing Systems 30: Annual Conference on Neural Information Processing Systems 2017, December 4-9, 2017, Long Beach, CA, {USA}}, pages 5998--6008.

\bibitem[{Vaudaux et~al.(2023)Vaudaux, Bazzoli, Coavoux, Vial, and Verg{\`e}s}]{vaudaux-etal-2023-pretrained}
Olivia Vaudaux, Caroline Bazzoli, Maximin Coavoux, G{\'e}raldine Vial, and {\'E}tienne Verg{\`e}s. 2023.
\newblock \href {https://doi.org/10.18653/v1/2023.nllp-1.5} {Pretrained language models v. court ruling predictions: A case study on a small dataset of {F}rench court of appeal rulings}.
\newblock In \emph{Proceedings of the Natural Legal Language Processing Workshop 2023}, pages 38--43, Singapore. Association for Computational Linguistics.

\bibitem[{Vieira et~al.(2021)Vieira, O’Hagan, and O’Sullivan}]{vieira2021understanding}
Lucas~Nunes Vieira, Minako O’Hagan, and Carol O’Sullivan. 2021.
\newblock Understanding the societal impacts of machine translation: a critical review of the literature on medical and legal use cases.
\newblock \emph{Information, Communication \& Society}, 24(11):1515--1532.

\bibitem[{Vilar et~al.(2023)Vilar, Freitag, Cherry, Luo, Ratnakar, and Foster}]{vilar2023prompting}
David Vilar, Markus Freitag, Colin Cherry, Jiaming Luo, Viresh Ratnakar, and George Foster. 2023.
\newblock Prompting palm for translation: Assessing strategies and performance.
\newblock In \emph{Proceedings of the 61st Annual Meeting of the Association for Computational Linguistics (Volume 1: Long Papers)}, pages 15406--15427.

\bibitem[{Wiesmann(2019)}]{wiesmann2019machine}
Eva Wiesmann. 2019.
\newblock {Machine translation in the field of law: A study of the translation of Italian legal texts into German}.
\newblock \emph{Comparative Legilinguistics}, 37(1):117--153.

\bibitem[{Yang et~al.(2024)Yang, Shen, Guo, Wang, Cao, Zhang, and Tao}]{yang2024modelmergingllmsmllms}
Enneng Yang, Li~Shen, Guibing Guo, Xingwei Wang, Xiaochun Cao, Jie Zhang, and Dacheng Tao. 2024.
\newblock \href {https://arxiv.org/abs/2408.07666} {Model merging in {LLMs, MLLMs}, and beyond: Methods, theories, applications and opportunities}.
\newblock \emph{Preprint}, arXiv:2408.07666.

\bibitem[{Zhang et~al.(2024)Zhang, Zhao, and Eger}]{zhang2024good}
Ran Zhang, Wei Zhao, and Steffen Eger. 2024.
\newblock {How Good Are LLMs for Literary Translation, Really? Literary Translation Evaluation with Humans and LLMs}.
\newblock \emph{arXiv preprint arXiv:2410.18697}.

\bibitem[{Zhang et~al.(2020)Zhang, Kishore, Wu, Weinberger, and Artzi}]{zhang_bertscore_2020}
Tianyi Zhang, Varsha Kishore, Felix Wu, Kilian~Q. Weinberger, and Yoav Artzi. 2020.
\newblock \href {http://arxiv.org/abs/1904.09675} {{BERTScore}: {Evaluating} {Text} {Generation} with {BERT}}.
\newblock \emph{arXiv:1904.09675 [cs]}.
\newblock ArXiv: 1904.09675.

\bibitem[{Zheng et~al.(2023)Zheng, Chiang, Sheng, Zhuang, Wu, Zhuang, Lin, Li, Li, Xing, Zhang, Gonzalez, and Stoica}]{zheng2023judgingllmasajudgemtbenchchatbot}
Lianmin Zheng, Wei-Lin Chiang, Ying Sheng, Siyuan Zhuang, Zhanghao Wu, Yonghao Zhuang, Zi~Lin, Zhuohan Li, Dacheng Li, Eric~P. Xing, Hao Zhang, Joseph~E. Gonzalez, and Ion Stoica. 2023.
\newblock \href {https://arxiv.org/abs/2306.05685} {Judging llm-as-a-judge with mt-bench and chatbot arena}.
\newblock \emph{Preprint}, arXiv:2306.05685.

\bibitem[{Zhong et~al.(2020)Zhong, Xiao, Tu, Zhang, Liu, and Sun}]{DBLP:conf/aaai/ZhongXTZ0S20}
Haoxi Zhong, Chaojun Xiao, Cunchao Tu, Tianyang Zhang, Zhiyuan Liu, and Maosong Sun. 2020.
\newblock \href {https://doi.org/10.1609/AAAI.V34I05.6519} {{JEC-QA:} {A} legal-domain question answering dataset}.
\newblock In \emph{The Thirty-Fourth {AAAI} Conference on Artificial Intelligence, {AAAI} 2020, The Thirty-Second Innovative Applications of Artificial Intelligence Conference, {IAAI} 2020, The Tenth {AAAI} Symposium on Educational Advances in Artificial Intelligence, {EAAI} 2020, New York, NY, USA, February 7-12, 2020}, pages 9701--9708. {AAAI} Press.

\bibitem[{Zhu et~al.(2024)Zhu, Liu, Dong, Xu, Huang, Kong, Chen, and Li}]{DBLP:conf/naacl/ZhuLDXHKCL24}
Wenhao Zhu, Hongyi Liu, Qingxiu Dong, Jingjing Xu, Shujian Huang, Lingpeng Kong, Jiajun Chen, and Lei Li. 2024.
\newblock \href {https://doi.org/10.18653/V1/2024.FINDINGS-NAACL.176} {Multilingual machine translation with large language models: Empirical results and analysis}.
\newblock In \emph{Findings of the Association for Computational Linguistics: {NAACL} 2024, Mexico City, Mexico, June 16-21, 2024}, pages 2765--2781. Association for Computational Linguistics.

\bibitem[{Ziemski et~al.(2016)Ziemski, Junczys-Dowmunt, and Pouliquen}]{un-parallel-corpus}
Micha{\l} Ziemski, Marcin Junczys-Dowmunt, and Bruno Pouliquen. 2016.
\newblock The {U}nited {N}ations parallel corpus {V}1. 0.
\newblock In \emph{Proceedings of the Tenth International Conference on Language Resources and Evaluation (LREC'16)}, pages 3530--3534.

\end{thebibliography}

\appendix
\onecolumn
\counterwithin{figure}{section}
\counterwithin{table}{section}

\section{Use of AI Assistants}
We used GPT-4o and Claude Sonnet 3.5 for coding, shortening texts and editing LaTeX more efficiently.

\section{Corpus Distribution of Text Lengths}
\label{sec:text-lengths}
\begin{figure*}[!htb]
    \centering
    \subtable[CH-Press-Trans dataset.]{
    \includegraphics[width=0.6\columnwidth]{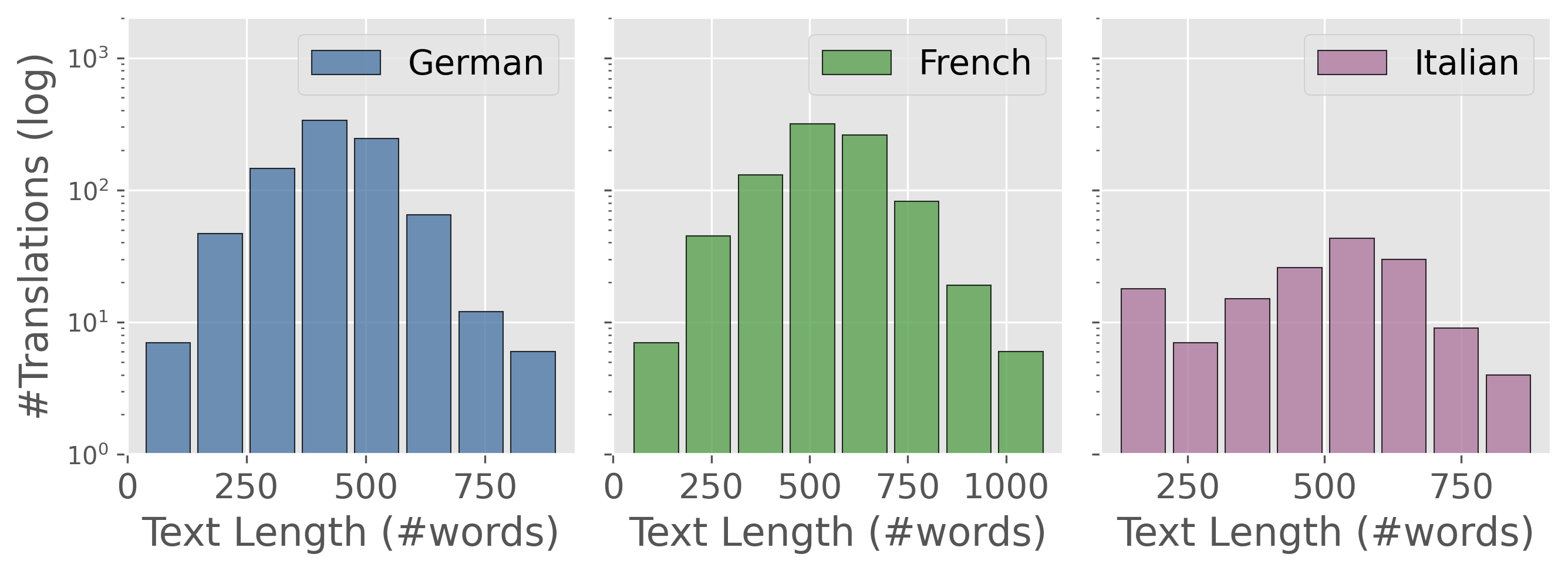}
    \label{fig:word-count-press}
    }
    \hspace{1.5ex}
    \subtable[CH-Headnote-Trans dataset (Text-Level).]{
    \includegraphics[width=0.6\columnwidth]{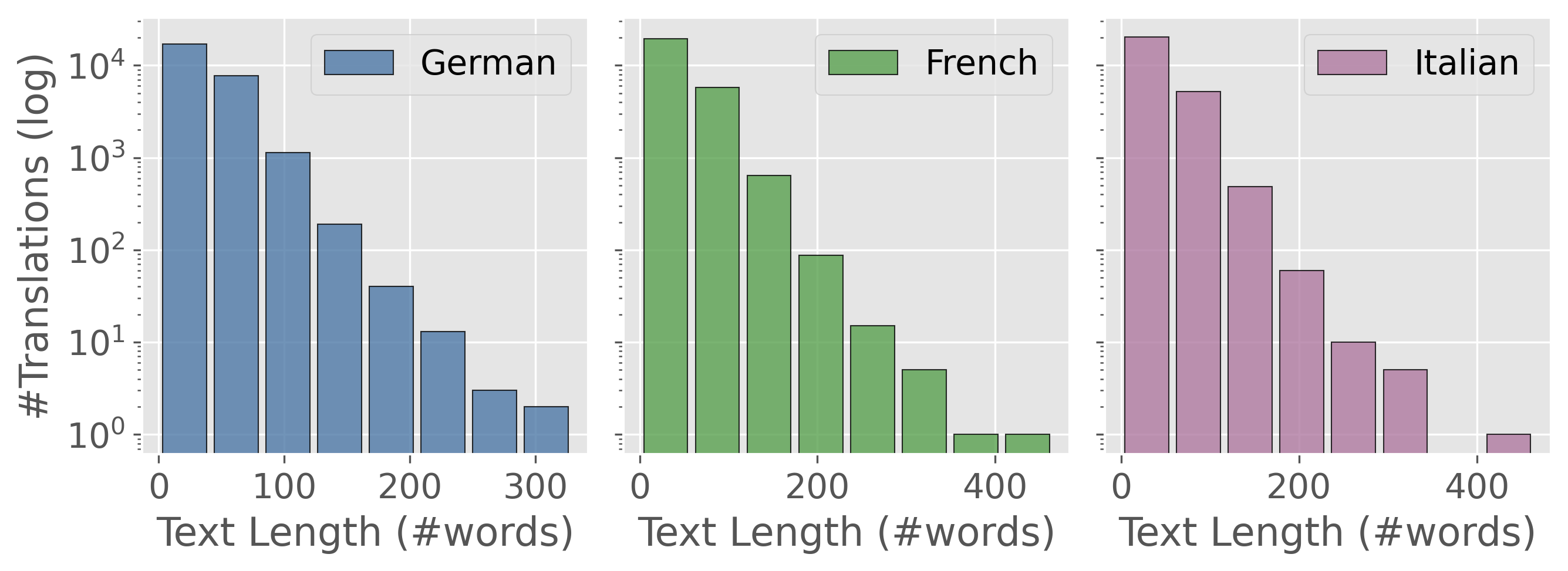}
    \label{fig:word-count-headnote-text}
    }
    \subtable[CH-Law-Trans dataset (Paragraph-Level).]{
    \includegraphics[width=\textwidth]{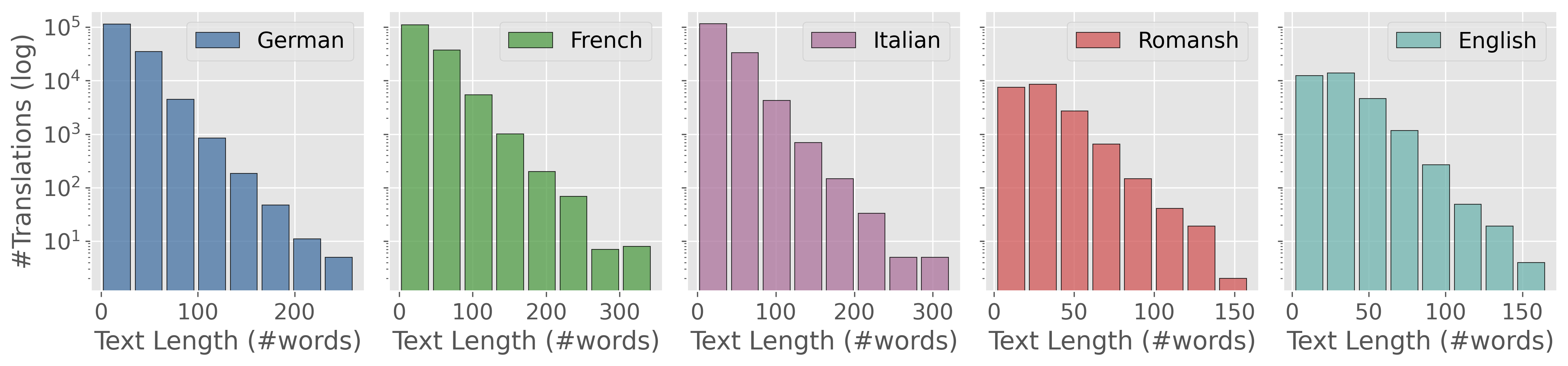}
    \label{fig:word-count-law-paragraph}
    }
    \vspace{-2ex}
    \caption{SwiLTra-Bench text length distribution (training set).}
    \label{fig:word-count}
    \vspace{-2ex}
\end{figure*}

\begin{figure*}[!htb]
    \centering
    \subtable[CH-Press-Trans dataset.]{
    \includegraphics[width=0.45\columnwidth]{figures/word-count-press.png}
    \label{fig:word-count-press}
    }
    \subtable[CH-Headnote-Trans dataset (BGE-Level).]{
    \includegraphics[width=0.45\columnwidth]{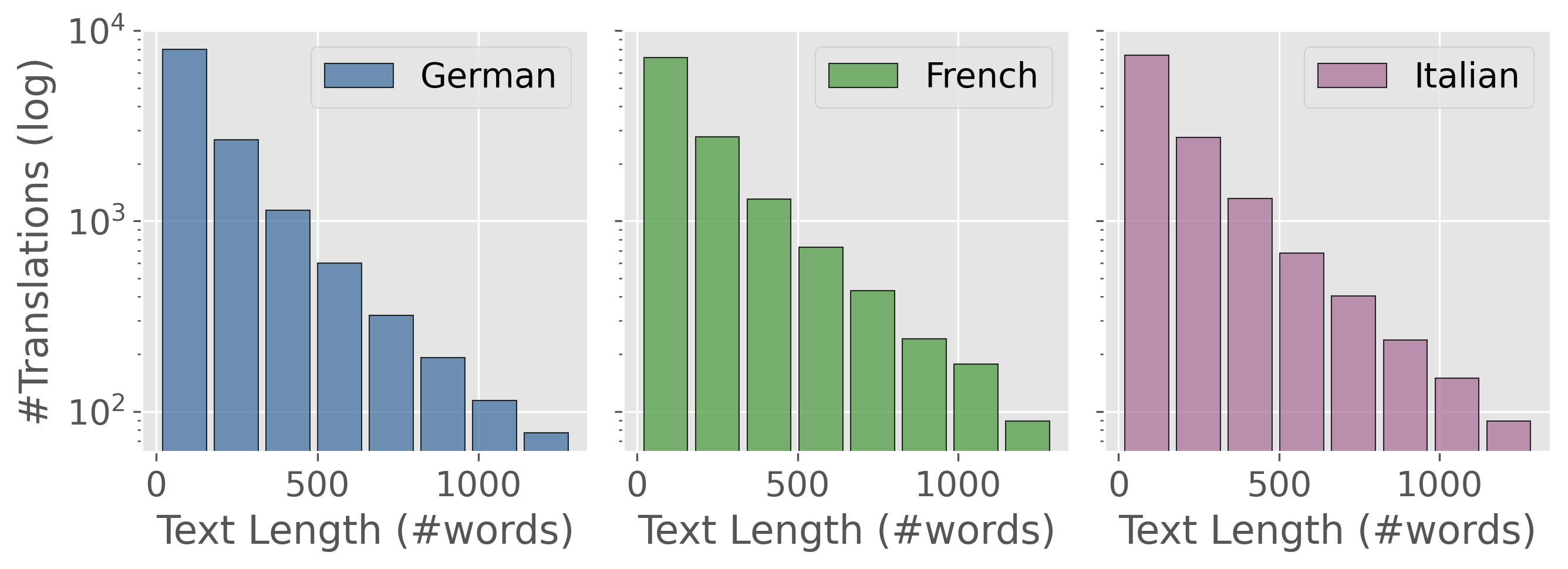}
    \label{fig:word-count-headnote-bge}
    }
    \subtable[CH-Headnote-Trans dataset (Regeste-Level).]{
    \includegraphics[width=0.45\columnwidth]{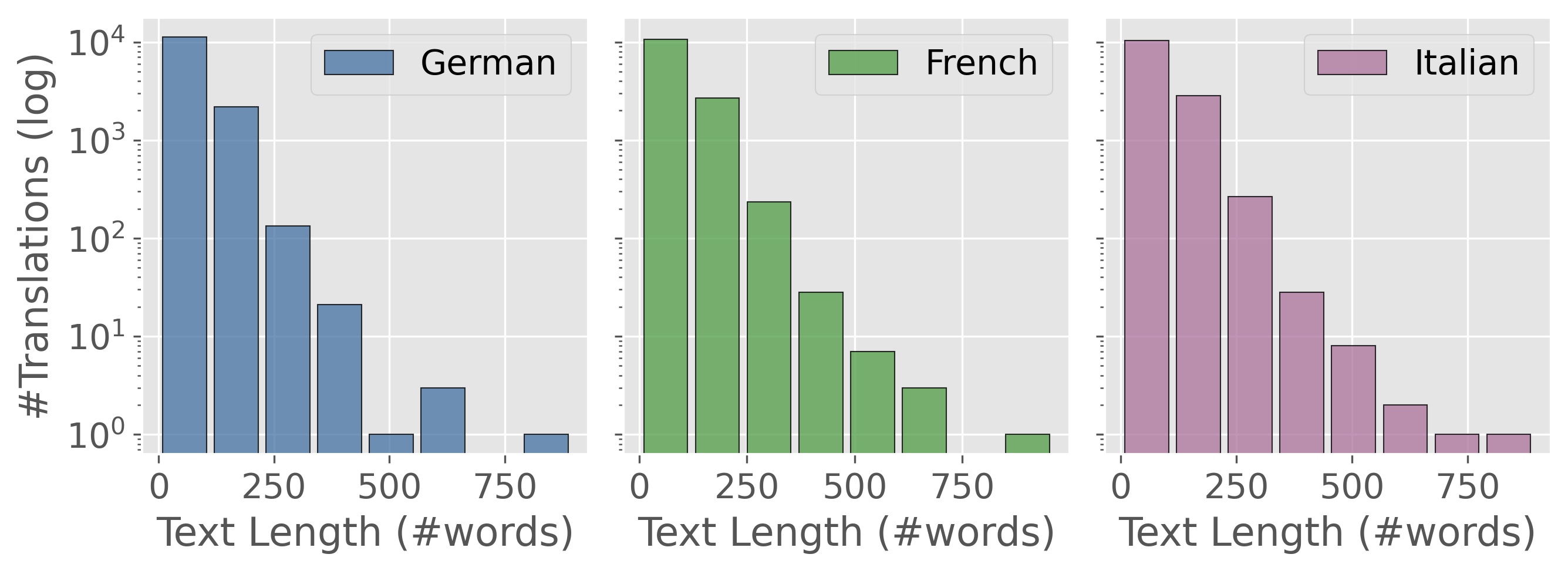}
    \label{fig:word-count-headnote-regeste}
    }
    \subtable[CH-Headnote-Trans dataset (Text-Level).]{
    \includegraphics[width=0.45\columnwidth]{figures/word-count-headnote-text.png}
    \label{fig:word-count-headnote-text}
    }
    \caption{Text length distribution of CH-Press-Trans and CH-Headnote-Trans dataset (training set).}
    \label{fig:word-count-headnote-press-app}
\end{figure*}
\begin{figure*}[htb]
    \centering
    \subtable[CH-Law-Trans dataset (Law-Level).]{
    \includegraphics[width=0.8\columnwidth]{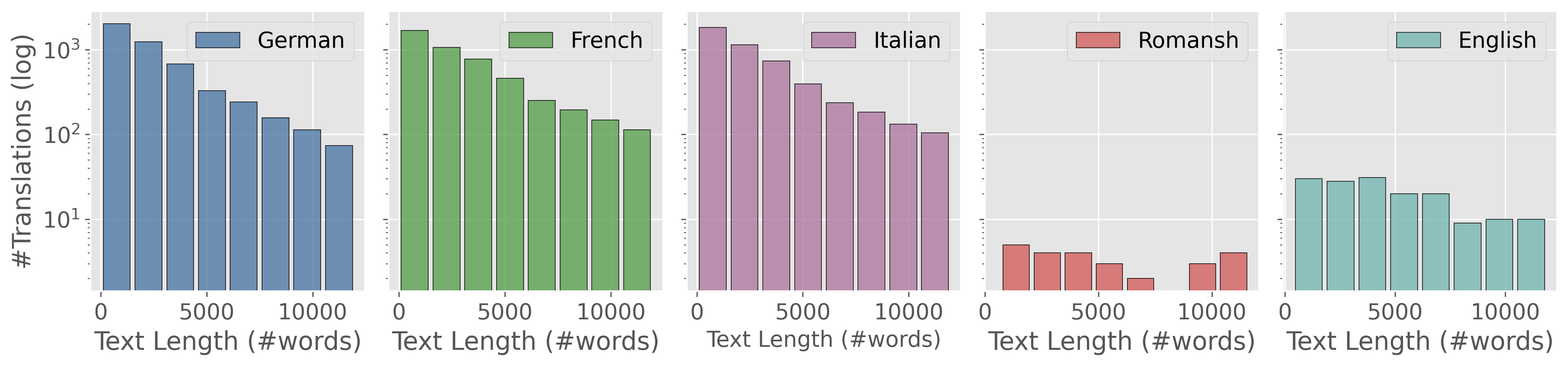}
    \label{fig:word-count-law-law}
    }
    \subtable[CH-Law-Trans dataset (Article-Level).]{
    \includegraphics[width=0.8\columnwidth]{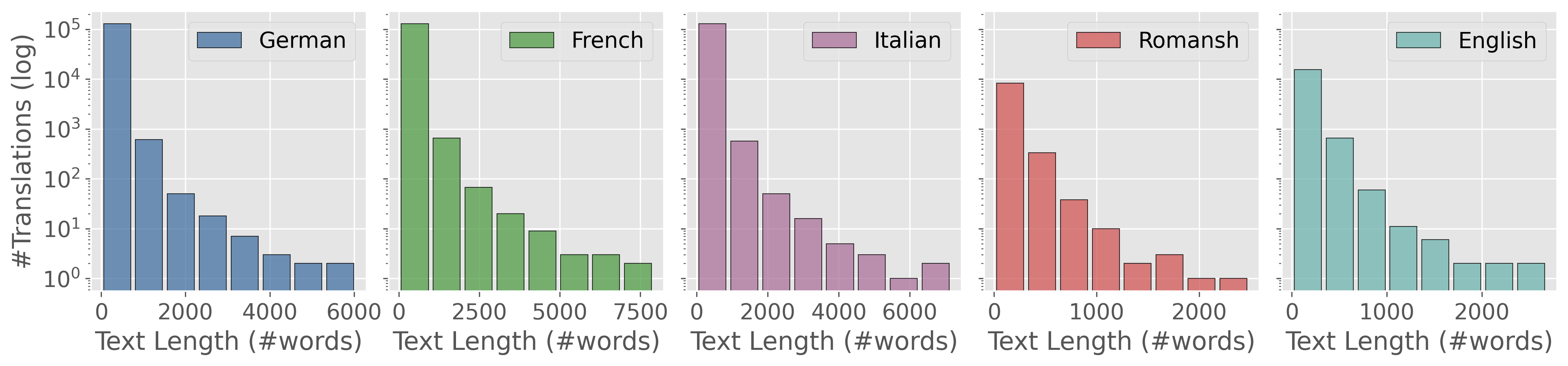}
    \label{fig:word-count-law-article}
    }
    \subtable[CH-Law-Trans dataset (Paragraph-Level).]{
    \includegraphics[width=0.8\columnwidth]{figures/word-count-law-paragraph.png}
    \label{fig:word-count-law-paragraph}
    }
    \caption{Text length distribution of CH-Law-Trans dataset (training set).}
    \label{fig:word-count-law-app}
\end{figure*}
\vspace{-2em}

\section{Additional Experimental Results}

\label{sec:additional-experiments}

\renewcommand{\arraystretch}{0.8}
\begin{table}[H]
\resizebox{\columnwidth}{!}{\begin{tabular}{llllrrrrr}
\toprule
\textbf{Model} & \textbf{Family} & \textbf{Category} & \textbf{Size} & \textbf{↑ GEMBA-MQM} & \textbf{↑ XCOMET} & \textbf{↑ BLEURT} & \textbf{↑ METEOR} & \textbf{↑ ChrF}\\
\midrule
Gemma-2-2B & Gemma & open & 2B & \cellcolor{teal!20}9.90 ± 0.1 & \cellcolor{teal!20}35.52 ± 0.1 & \cellcolor{teal!20}-102.08 ± 0.3 & \cellcolor{teal!20}6.97 ± 0.1 & \cellcolor{teal!26}11.12 ± 0.1 \\
SLT-Gemma-2-2B & Gemma & finetuned & 2B & \cellcolor{teal!63}72.71 ± 0.2 & \cellcolor{teal!64}82.39 ± 0.1 & \cellcolor{teal!67}26.72 ± 0.3 & \cellcolor{teal!66}61.52 ± 0.1 & \cellcolor{teal!69}79.48 ± 0.1 \\
Gemma-2-9B & Gemma & open & 9B & \cellcolor{teal!21}12.15 ± 0.1 & \cellcolor{teal!20}36.54 ± 0.1 & \cellcolor{teal!20}-102.19 ± 0.3 & \cellcolor{teal!20}7.48 ± 0.1 & \cellcolor{teal!20}0.00 ± 0.1 \\
SLT-Gemma-2-9B & Gemma & finetuned & 9B & \cellcolor{teal!69}82.54 ± 0.1 & \underline{\cellcolor{teal!69}87.62 ± 0.1} & \underline{\cellcolor{teal!69}32.89 ± 0.2} & \textbf{\cellcolor{teal!70}65.16 ± 0.1} & \cellcolor{teal!68}78.95 ± 0.1 \\
Llama-3.2-1B & Llama & open & 1B & \cellcolor{teal!31}27.23 ± 0.2 & \cellcolor{teal!32}48.43 ± 0.2 & \cellcolor{teal!51}-15.64 ± 0.2 & \cellcolor{teal!39}29.70 ± 0.1 & \cellcolor{teal!44}39.87 ± 0.1 \\
SLT-Llama-3.2-1B & Llama & finetuned & 1B & \cellcolor{teal!57}64.14 ± 0.2 & \cellcolor{teal!59}76.40 ± 0.1 & \cellcolor{teal!67}26.35 ± 0.2 & \cellcolor{teal!64}59.03 ± 0.1 & \cellcolor{teal!69}79.76 ± 0.1 \\
Llama-3.2-3B & Llama & open & 3B & \cellcolor{teal!50}54.13 ± 0.2 & \cellcolor{teal!50}67.43 ± 0.2 & \cellcolor{teal!57}0.59 ± 0.2 & \cellcolor{teal!47}38.57 ± 0.1 & \cellcolor{teal!51}50.47 ± 0.1 \\
SLT-Llama-3.2-3B & Llama & finetuned & 3B & \cellcolor{teal!65}75.56 ± 0.2 & \cellcolor{teal!65}83.39 ± 0.1 & \cellcolor{teal!68}30.47 ± 0.2 & \cellcolor{teal!67}62.54 ± 0.1 & \cellcolor{teal!69}79.32 ± 0.1 \\
Llama-3.1-8B & Llama & open & 8B & \cellcolor{teal!59}67.09 ± 0.2 & \cellcolor{teal!57}75.03 ± 0.2 & \cellcolor{teal!60}6.25 ± 0.2 & \cellcolor{teal!51}43.72 ± 0.1 & \cellcolor{teal!51}50.56 ± 0.1 \\
SLT-Llama-3.1-8B & Llama & finetuned & 8B & \cellcolor{teal!68}80.23 ± 0.1 & \cellcolor{teal!68}86.04 ± 0.1 & \cellcolor{teal!69}31.91 ± 0.2 & \cellcolor{teal!69}64.17 ± 0.1 & \textbf{\cellcolor{teal!70}80.89 ± 0.1} \\
Phi-3.5-mini & Phi & open & 3.8B & \cellcolor{teal!25}17.96 ± 0.2 & \cellcolor{teal!26}41.93 ± 0.1 & \cellcolor{teal!23}-92.40 ± 0.3 & \cellcolor{teal!22}9.66 ± 0.1 & \cellcolor{teal!27}11.42 ± 0.1 \\
SLT-Phi-3.5-mini & Phi & finetuned & 3.8B & \cellcolor{teal!63}73.90 ± 0.2 & \cellcolor{teal!62}80.31 ± 0.1 & \cellcolor{teal!61}10.33 ± 0.2 & \cellcolor{teal!62}56.75 ± 0.1 & \cellcolor{teal!67}76.72 ± 0.1 \\
Phi-3-medium & Phi & open & 14B & \cellcolor{teal!27}21.33 ± 0.2 & \cellcolor{teal!23}38.91 ± 0.1 & \cellcolor{teal!27}-81.04 ± 0.3 & \cellcolor{teal!25}13.45 ± 0.1 & \cellcolor{teal!30}17.74 ± 0.1 \\
SLT-Phi-3-medium & Phi & finetuned & 14B & \cellcolor{teal!69}81.56 ± 0.1 & \cellcolor{teal!69}87.38 ± 0.1 & \cellcolor{teal!69}32.39 ± 0.2 & \cellcolor{teal!69}64.16 ± 0.1 & \underline{\cellcolor{teal!69}80.40 ± 0.1} \\
Qwen2.5-0.5B & Qwen & open & 0.5B & \cellcolor{teal!20}9.82 ± 0.2 & \cellcolor{teal!25}41.36 ± 0.2 & \cellcolor{teal!34}-61.96 ± 0.2 & \cellcolor{teal!26}14.53 ± 0.1 & \cellcolor{teal!37}28.21 ± 0.1 \\
SLT-Qwen2.5-0.5B & Qwen & finetuned & 0.5B & \cellcolor{teal!49}52.37 ± 0.2 & \cellcolor{teal!52}69.48 ± 0.2 & \cellcolor{teal!66}22.67 ± 0.2 & \cellcolor{teal!62}56.77 ± 0.1 & \cellcolor{teal!68}78.66 ± 0.1 \\
Qwen2.5-1.5B & Qwen & open & 1.5B & \cellcolor{teal!37}35.21 ± 0.2 & \cellcolor{teal!41}58.12 ± 0.2 & \cellcolor{teal!40}-46.89 ± 0.3 & \cellcolor{teal!33}22.55 ± 0.1 & \cellcolor{teal!42}36.26 ± 0.1 \\
SLT-Qwen2.5-1.5B & Qwen & finetuned & 1.5B & \cellcolor{teal!60}69.58 ± 0.2 & \cellcolor{teal!62}80.43 ± 0.1 & \cellcolor{teal!67}26.90 ± 0.2 & \cellcolor{teal!65}60.45 ± 0.1 & \cellcolor{teal!68}78.13 ± 0.1 \\
Qwen2.5-3B & Qwen & open & 3B & \cellcolor{teal!46}48.85 ± 0.2 & \cellcolor{teal!44}61.18 ± 0.2 & \cellcolor{teal!53}-11.88 ± 0.2 & \cellcolor{teal!43}34.77 ± 0.1 & \cellcolor{teal!47}44.46 ± 0.1 \\
SLT-Qwen2.5-3B & Qwen & finetuned & 3B & \cellcolor{teal!64}74.33 ± 0.2 & \cellcolor{teal!65}82.99 ± 0.1 & \cellcolor{teal!67}27.78 ± 0.2 & \cellcolor{teal!66}61.61 ± 0.1 & \cellcolor{teal!68}78.03 ± 0.1 \\
Qwen2.5-7B & Qwen & open & 7B & \cellcolor{teal!53}58.79 ± 0.2 & \cellcolor{teal!52}69.07 ± 0.2 & \cellcolor{teal!58}0.73 ± 0.2 & \cellcolor{teal!47}39.41 ± 0.1 & \cellcolor{teal!46}42.67 ± 0.1 \\
SLT-Qwen2.5-7B & Qwen & finetuned & 7B & \cellcolor{teal!67}78.96 ± 0.1 & \cellcolor{teal!68}86.03 ± 0.1 & \cellcolor{teal!69}31.07 ± 0.2 & \cellcolor{teal!68}63.40 ± 0.1 & \cellcolor{teal!67}77.54 ± 0.1 \\
Qwen2.5-14B & Qwen & open & 14B & \cellcolor{teal!63}72.70 ± 0.2 & \cellcolor{teal!62}79.54 ± 0.1 & \cellcolor{teal!61}9.27 ± 0.2 & \cellcolor{teal!52}45.06 ± 0.1 & \cellcolor{teal!54}56.13 ± 0.1 \\
SLT-Qwen2.5-14B & Qwen & finetuned & 14B & \textbf{\cellcolor{teal!70}82.78 ± 0.1} & \cellcolor{teal!69}87.46 ± 0.1 & \cellcolor{teal!69}32.37 ± 0.2 & \cellcolor{teal!69}64.73 ± 0.1 & \cellcolor{teal!68}78.56 ± 0.1 \\
Qwen2.5-32B & Qwen & open & 32B & \cellcolor{teal!61}70.30 ± 0.2 & \cellcolor{teal!58}76.34 ± 0.1 & \cellcolor{teal!60}6.91 ± 0.2 & \cellcolor{teal!52}45.33 ± 0.1 & \cellcolor{teal!55}57.94 ± 0.1 \\
SLT-Qwen2.5-32B & Qwen & finetuned & 32B & \underline{\cellcolor{teal!69}82.77 ± 0.1} & \textbf{\cellcolor{teal!70}87.90 ± 0.1} & \textbf{\cellcolor{teal!70}33.20 ± 0.2} & \underline{\cellcolor{teal!69}65.07 ± 0.1} & \cellcolor{teal!67}76.75 ± 0.1 \\
\bottomrule
\end{tabular}}
\caption{Base models and their finetuned versions across different families and sizes.}
\label{tab:finetune-vs-base}
\end{table}

\newpage
\section{Corpus Examples}
\label{sec:corpus-examples}

\begin{table*}[!htb]
    \centering
    \resizebox{\textwidth}{!}{
    \begin{tabular}{cll}
    \toprule
    Dataset & Field & Comment \\
    \midrule
    \multirow{5}{*}{\small \rotatebox{90}{CH-Press-Trans}} & \texttt{filename} & Unique identifier of each press release \\ 
    & \texttt{de\_text} & Press release content in German \\
    & \texttt{fr\_text} & Press release content in French \\
    & \texttt{it\_text} & Press release content in Italian \\
    & \texttt{has\_all\_langs} & Binary indicator of language availablity \\
    \midrule
    \multirow{14}{*}{\rotatebox{90}{CH-Law-Trans}} & \texttt{abbreviation} & The abbreviation of the law \\
    & \texttt{url} & URL linking to the legal text on Fedlex \\
    & \texttt{rsNr} & Swiss federal register number \\
    & \texttt{artNr} & Article number \\
    & \texttt{parNr} & Paragraph number \\
    & \texttt{dateApplicability} & Date of applicability of the law \\
    & \texttt{\{de/fr/it/rm/en\}\_lawTitle} & Law titles in different languages \\
    & \texttt{\{de/fr/it/rm/en\}\_artTitle} & Article titles in different languages \\
    & \texttt{\{de/fr/it/rm/en\}\_lawText} & Full law texts in different languages \\
    & \texttt{\{de/fr/it/rm/en\}\_artText} & Full article texts in different languages \\
    & \texttt{\{de/fr/it/rm/en\}\_parText} & Full paragraph texts in different languages \\
    & \texttt{\{de/fr/it/rm/en\}\_lawHtml} & Law texts in HTML format in different languages \\
    & \texttt{\{de/fr/it/rm/en\}\_artHtml} & Article texts in HTML format in different languages \\
    & \texttt{\{de/fr/it/rm/en\}\_parHtml} & Paragraph texts in HTML format in different languages \\
    \midrule
    \multirow{10}{*}{\rotatebox{90}{CH-Headnote-Trans}} & \texttt{bge} & Case identifier \\
    & \texttt{year} & Year of the court decision \\
    & \texttt{volume} & Volume number of the court decision \\
    & \texttt{pageNumber} & Page number of the court decision \\
    & \texttt{regesteNumber} & Number assigned to the regeste \\
    & \texttt{textNumber} & Number assigned to the specific text extract \\
    & \texttt{\{de/fr/it\}\_bgeText} & Full summary texts in different languages \\
    & \texttt{\{de/fr/it\}\_regesteText} & Regeste texts in different languages \\
    & \texttt{\{de/fr/it\}\_regesteTitle} & Regeste title in different languages \\
    & \texttt{\{de/fr/it\}\_text} & Text extract in different languages \\
    \bottomrule
    \end{tabular}
    }
    \caption{Structure of the three SwiLTra-Bench datasets. Parallel translations for Romansh and English are only available in parts of the CH-Law-Trans dataset.}
\end{table*}
\begin{listing*}[!htb]
\begin{lstlisting}
{     
    'de_abbreviation': BV,
    'de_artText': Das Schweizervolk und die Kantone Zürich, Bern, Luzern, Uri, Schwyz, Obwalden und Nidwalden, Glarus, Zug, Freiburg, Solothurn, Basel-Stadt und Basel-Landschaft, Schaffhausen, Appenzell Ausserrhoden und Appenzell Innerrhoden, St. Gallen, Graubünden, Aargau, Thurgau, Tessin, Waadt, Wallis, Neuenburg, Genf und Jura bilden die Schweizerische Eidgenossenschaft.,
    ...
    'de_artTitle': Art. 1 Schweizerische Eidgenossenschaft,
    
    'fr_abbreviation': Cst.,
    'fr_artText': Le peuple suisse et les cantons de Zurich, de Berne, de Lucerne, d'Uri, de Schwyz, d'Obwald et de Nidwald, de Glaris, de Zoug, de Fribourg, de Soleure, de Bâle-Ville et de Bâle-Campagne, de Schaffhouse, d'Appenzell Rhodes-Extérieures et d'Appenzell Rhodes-Intérieures, de Saint-Gall, des Grisons, d'Argovie, de Thurgovie, du Tessin, de Vaud, du Valais, de Neuchâtel, de Genève et du Jura forment la Confédération suisse.
    ...
    'fr_artTitle': Art. 1 Confédération suisse,

    'it_abbreviation': Cost.,
    'it_artText': Il Popolo svizzero e i Cantoni di Zurigo, Berna, Lucerna, Uri, Svitto, Obvaldo e Nidvaldo, Glarona, Zugo, Friburgo, Soletta, Basilea Città e Basilea Campagna, Sciaffusa, Appenzello Esterno e Appenzello Interno, San Gallo, Grigioni, Argovia, Turgovia, Ticino, Vaud, Vallese, Neuchâtel, Ginevra e Giura costituiscono la Confederazione Svizzera.
    ...
    'it_artTitle': Art. 1 Confederazione Svizzera,


    'rm_abbreviation': Cst., 
    'rm_artText': Il pievel svizzer ed ils chantuns Turitg, Berna, Lucerna, Uri, Sviz, Sursilvania e Sutsilvania, Glaruna, Zug, Friburg, Soloturn, Basilea-Citad e Basilea-Champagna, Schaffusa, Appenzell Dadens ed Appenzell Dador, Son Gagl, Grischun, Argovia, Turgovia, Tessin, Vad, Vallais, Neuchâtel, Genevra e Giura furman la Confederaziun svizra.,
    ...
    'rm_artTitle': Art. 1 Confederaziun svizra, 

    'en_abbreviation': Cst.,
    'en_artText': The People and the Cantons of Zurich, Bern, Lucerne, Uri, Schwyz, Obwalden and Nidwalden, Glarus, Zug, Fribourg, Solothurn, Basel Stadt and Basel Landschaft, Schaffhausen, Appenzell Ausserrhoden and Appenzell Innerrhoden, St. Gallen, Graubünden, Aargau, Thurgau, Ticino, Vaud, Valais, Neuchâtel, Geneva, and Jura form the Swiss Confederation., 
    ...
    'en_artTitle': Art. 1 The Swiss Confederation,
}
\end{lstlisting}
\caption{An Example of CH-Law-Trans: Article Dataset} 
\label{lst:example-law}
\end{listing*}

\begin{listing*}[!htb]
\begin{lstlisting}
{     
    'bge': 100-IA-231,
    'year': 100,
    'volume': IA,
    'pageNumber': 231,
    
    'de_bgeText': Art. 85 lit. a OG. Ungültigerklärung einer kommunalen Volksinitiative wegen materieller Unvereinbarkeit mit dem kantonalen Recht. 1. Wieweit muss die Behörde beim Entscheid über die Gültigkeit einer kommunalen Initiative berücksichtigen, dass deren materielle Widerrechtlichkeit durch Annahme eines gleichzeitig eingereichten kantonalen Volksbegehrens dahinfallen könnte? (Erw. 2). 2. Die Verkehrsbetriebe der Stadt Zürich sind eine zur Eigenwirtschaftlichkeit verpflichtete ''produktive Unternehmun'' im Sinne von par. 129 des kantonalen Gemeindegesetzes. Die stadtzürcherische ''Gratistram-Initiative'', mit welcher ein grundsätzlicher Verzicht auf die Erhebung von Benützungsgebühren gefordert wurde, durfte daher wegen Unvereinbarkeit mit dem kantonalen Recht für ungültig erklärt werden (Erw. 3).,
    
    'fr_bgeText': Art. 85 lit. a OJ. Décision niant la validité d'une initiative communale en raison de son incompatibilité matérielle avec le droit cantonal. 1. Dans quelle mesure l'autorité qui se prononce sur la validité d'une initiative communale doit-elle tenir compte du fait que le contenu de cette dernière, contraire au droit, pourrait ne plus l'être en raison de l'acceptation d'une initiative cantonale déposée simultanément? (consid. 2). 2. Les entreprises de transport de la ville de Zurich, qui doivent être gérées selon les principes de l'économie industrielle, sont une ''entreprise à caractère productif'' au sens de l'art. 129 de la loi cantonale sur les communes. L'initiative communale zurichoise ''Gratistram'', qui exigeait en principe la suppression de toute taxe d'utilisation, pouvait être déclarée non valable en raison de son incompatibilité avec le droit cantonal (consid. 3).,
    
    'it_bgeText': Art. 85 lett. a OG. Diniego della validità di un'iniziativa comunale a causa della sua incompatibilità con il diritto cantonale. 1. In quale misura l'autorità che si pronuncia sulla validità di una iniziativa comunale deve tener conto del fatto che il contenuto di quest'ultima, contrario alla legge, cesserebbe d\'esserlo ove fosse accettata una iniziativa cantonale presentata nello stesso tempo? (consid. 2). 2. Le imprese di trasporto della città di Zurigo costituiscono una ''azienda produttiva'' ai sensi dell'art. 129 della legge cantonale sui comuni, tenuta come tale ad un esercizio secondo criteri economici. L'iniziativa comunale zurighese per il tram gratuito, che esigeva in linea di principio la soppressione d'ogni tassa d'utilizzazione, poteva quindi essere dichiarata invalida per la sua incompatibilità con il diritto cantonale (consid. 3).
}
\end{lstlisting}
\caption{An Example of CH-Headnote-Trans: BGE Dataset} 
\label{lst:example-headnote}
\end{listing*}
\begin{listing*}[!htb]
\begin{lstlisting}
{     
    'de_text': ... Das BJ wies zuerst das Gesuch und dann die gegen diese Verfügung erhobene Einsprache des Betroffenen ab. Das Bundesverwaltungsgericht hiess die Beschwerde des Betroffenen gut, hob den Einspracheentscheid des BJ auf und wies die Angelegenheit dem BJ zurück, wogegen das BJ beim Bundesgericht eine Beschwerde eingereicht hat.

    Das Bundesgericht weist die Beschwerde ab. Gestützt auf eine vertiefte Auslegung des AFZFG kommt das Bundesgericht zum Schluss, dass ein Kind auch nach einer Adoption durch seine vormaligen Pflegeeltern als fremdplatziert im Sinne von Artikel 2 Buchstabe b des AFZFG gilt, womit es auch nach der Adoption von einer Fremdplatzierung betroffen ist und die Opfereigenschaft nach Artikel 2 Buchstabe d AFZFG erfüllen kann.

    'fr_text': ... L'OFJ a rejeté tant la demande que l'opposition formées par l'intéressé. Le Tribunal administratif fédéral a admis le recours de l'intéressé, annulé la décision sur opposition de l'OFJ et renvoyé l'affaire à l'OFJ, lequel a déposé un recours auprès du Tribunal fédéral.

    Le Tribunal fédéral rejette le recours. Sur la base d'une interprétation approfondie de la LMCFA, il parvient à la conclusion qu'un enfant doit être considéré comme ayant fait l'objet d'un placement extrafamilial au sens de l'article 2 lettre b LMCFA même après avoir été adopté par ses parents nourriciers, si bien que la qualité de personne concernée et le statut de victime aux termes de l'article 2 lettre d LMCFA doivent lui être reconnus même après l'adoption.

    'it_text': ... L'UFG ha respinto prima la domanda e poi l'opposizione interposta dall'interessato contro questa decisione. Il Tribunale amministrativo federale ha accolto il ricorso dell'interessato, ha annullato la decisione su opposizione resa dall'UFG e ha rinviato la questione all'UFG, che ha presentato ricorso al Tribunale federale.

    Il Tribunale federale respinge il ricorso. Sulla base di un'interpretazione approfondita della LMCCE, il Tribunale federale giunge alla conclusione che si deve ritenere che un bambino ha subito un collocamento extrafamiliare ai sensi dell'articolo 2 lettera b LMCCE anche dopo essere stato adottato dai genitori affilianti ed è pertanto riconosciuto come persona oggetto di misure nonché vittima secondo l'articolo 2 lettera d LMCCE anche dopo l'adozione.
}
\end{lstlisting}
\caption{An Example of CH-Press-Trans Dataset} 
\label{lst:example-press}
\end{listing*}

\onecolumn

\newpage
\section{Judge Correlations}
\label{sec:judge-correlations}
\begin{table}[!htb]
\resizebox{\columnwidth}{!}{\begin{tabular}{lllll}
\toprule
\textbf{Metric} & \textbf{Spearman (Bootstrap)} & \textbf{Spearman (CV)} & \textbf{RMSE (CV)} & \textbf{MAE (CV)}\\
\midrule
judge-ensemble & 0.536 [0.453, 0.608] & 0.533 ± 0.080 & 16.090 ± 1.424 & 12.979 ± 0.882 \\
gemini-1-5-flash-codebook-diverse-deduction & 0.506 [0.424, 0.586] & 0.504 ± 0.074 & 15.215 ± 2.216 & 11.100 ± 0.670 \\
XCOMET-XXL & 0.484 [0.403, 0.567] & 0.477 ± 0.093 & 14.877 ± 1.372 & 10.204 ± 0.748 \\
gpt-4o-mini-codebook-single-deduction & 0.474 [0.387, 0.551] & 0.477 ± 0.095 & 22.168 ± 3.064 & 16.944 ± 1.691 \\
gemini-1-5-flash-codebook-single-deduction & 0.461 [0.374, 0.542] & 0.466 ± 0.069 & 16.049 ± 2.212 & 10.990 ± 0.720 \\
gpt-4o-mini-codebook-diverse-deduction & 0.459 [0.378, 0.538] & 0.459 ± 0.094 & 22.527 ± 2.678 & 17.138 ± 1.113 \\
gpt-4o-codebook-single-deduction & 0.444 [0.352, 0.538] & 0.447 ± 0.070 & 29.020 ± 3.982 & 19.606 ± 1.951 \\
gpt-4o-codebook-diverse-deduction & 0.427 [0.334, 0.516] & 0.412 ± 0.044 & 30.902 ± 1.843 & 21.537 ± 0.555 \\
gpt-4o-detailed-single-absolute & 0.418 [0.330, 0.501] & 0.427 ± 0.052 & 35.940 ± 4.207 & 24.286 ± 3.382 \\
gpt-4o-mini-basic-single-absolute & 0.413 [0.320, 0.497] & 0.422 ± 0.067 & 26.736 ± 1.817 & 21.370 ± 2.036 \\
gpt-4o-basic-single-absolute & 0.411 [0.320, 0.499] & 0.411 ± 0.131 & 20.655 ± 2.947 & 14.596 ± 1.976 \\
gpt-4o-detailed-diverse-absolute & 0.378 [0.284, 0.464] & 0.380 ± 0.090 & 34.786 ± 3.711 & 22.995 ± 3.173 \\
gpt-4o-mini-detailed-single-absolute & 0.378 [0.282, 0.471] & 0.384 ± 0.069 & 35.437 ± 1.866 & 30.302 ± 1.718 \\
gpt-4o-basic-diverse-absolute & 0.376 [0.281, 0.467] & 0.383 ± 0.087 & 21.780 ± 2.249 & 13.550 ± 1.382 \\
gpt-4o-mini-detailed-diverse-absolute & 0.358 [0.259, 0.450] & 0.364 ± 0.097 & 36.480 ± 2.365 & 30.828 ± 2.387 \\
bleurt\_large & 0.356 [0.261, 0.445] & 0.364 ± 0.147 & 63.110 ± 5.225 & 58.102 ± 5.064 \\
gpt-4o-mini-basic-diverse-absolute & 0.354 [0.263, 0.447] & 0.361 ± 0.048 & 28.363 ± 1.315 & 22.393 ± 1.347 \\
gemini-1-5-pro-basic-single-absolute & 0.306 [0.209, 0.404] & 0.295 ± 0.083 & 36.203 ± 3.618 & 22.993 ± 3.010 \\
gemini-1-5-pro-codebook-diverse-deduction & 0.303 [0.200, 0.397] & 0.298 ± 0.095 & 34.528 ± 3.882 & 20.580 ± 2.477 \\
GEMBA-MQM\_gpt-4o & 0.293 [0.189, 0.383] & 0.289 ± 0.093 & 18.331 ± 1.743 & 12.698 ± 0.787 \\
gemini-1-5-pro-codebook-single-deduction & 0.292 [0.195, 0.392] & 0.292 ± 0.074 & 36.718 ± 2.663 & 21.816 ± 2.158 \\
gemini-1-5-flash-detailed-single-absolute & 0.281 [0.190, 0.375] & 0.275 ± 0.049 & 29.067 ± 3.070 & 18.709 ± 1.651 \\
gemini-1-5-flash-basic-single-absolute & 0.270 [0.179, 0.359] & 0.279 ± 0.110 & 33.283 ± 5.239 & 20.412 ± 3.287 \\
gemini-1-5-flash-basic-diverse-absolute & 0.255 [0.157, 0.351] & 0.249 ± 0.069 & 27.649 ± 5.852 & 16.756 ± 3.233 \\
gemini-1-5-pro-basic-diverse-absolute & 0.252 [0.153, 0.351] & 0.250 ± 0.082 & 36.544 ± 3.470 & 22.990 ± 2.684 \\
gemini-1-5-pro-detailed-diverse-absolute & 0.247 [0.147, 0.345] & 0.250 ± 0.097 & 38.316 ± 2.063 & 26.098 ± 1.148 \\
gemini-1-5-pro-detailed-single-absolute & 0.235 [0.137, 0.334] & 0.244 ± 0.079 & 38.618 ± 2.362 & 27.798 ± 1.977 \\
gemini-1-5-flash-detailed-diverse-absolute & 0.231 [0.130, 0.328] & 0.225 ± 0.055 & 30.091 ± 4.891 & 19.671 ± 3.346 \\
BERTScore-F & 0.163 [0.066, 0.268] & 0.170 ± 0.053 & 36.723 ± 1.340 & 31.523 ± 1.772 \\
meteor & 0.160 [0.057, 0.259] & 0.164 ± 0.125 & 34.170 ± 3.444 & 29.270 ± 3.191 \\
\bottomrule
\end{tabular}}
\caption{Correlation metrics with human scores (with 95\% CIs and Cross-Validation)}
\label{tab:correlation-metrics-comprehensive}
\end{table}

\section{Judge Prompts}
\label{sec:judge-prompts}
\begin{listing*}[h]

\textbf{System Prompt}
\begin{lstlisting}
Act as a Judge specializing in the evaluation of translations of Swiss legal documents. Your task is to assess the accuracy, clarity, and fidelity of the model's translation to the golden translation, while considering the nuances of legal language.
\end{lstlisting}

\textbf{User Prompt}
\begin{lstlisting}
You will be provided with a source text, its golden translation, and the model's translation. Your task is to judge how correct the model's translation is based on the golden translation, and then give a correctness score. The correctness score should be one of the below numbers: 0.0 (totally wrong), 0.1, 0.2, 0.3, 0.4, 0.5, 0.6, 0.7, 0.8, 0.9, or 1.0 (totally right). You should first briefly give your reasoning process regarding how the model's translation conforms to or contradicts the golden translation, and then give the correctness score. The correctness score must strictly follow this format: \"[[score]]\", e.g., \"The correctness score: [[0.5]]\". Below are some examples.
\end{lstlisting}
\caption{The system and user prompt of the \textit{basic} judge setup.} 
\label{tab:judge-prompt-basic}
\end{listing*}

\begin{listing*}[h]

\textbf{System Prompt}
\begin{lstlisting}
You are a senior legal translator and quality assurance specialist with over 20 years of experience in Swiss law, certified by the Swiss Sworn Translators Association (Association suisse des traducteurs-jurés, ASTJ). You possess native-level proficiency in all Swiss national languages (German, French, Italian, and Romansh) as well as English, enabling precise evaluation of legal nuances across all linguistic combinations. Your task is to evaluate machine-translated legal texts for accuracy, clarity and fidelity to Swiss legal standards analyzing the subtle complexities of legal language. You excel at identifying even minor discrepancies and calibrating evaluation scores appropriately to reflect the severity of each error.
\end{lstlisting}

\textbf{User Prompt}
\begin{lstlisting}
INPUT FORMAT:
Source Text: [Original text in source language]
Golden Translation: [Reference professional translation]
Model Translation: [Machine-generated translation to be evaluated]

EVALUATION DIMENSIONS:
Accuracy: Semantic equivalence, correct legal terminology, and preservation of legal meaning.
Clarity: Logical flow, appropriate legal register, and unambiguous expression.
Fidelity: Adherence to Swiss legal conventions, jurisdiction-specific terminology, and formal register.

SCORING RUBRIC:
1.0: Perfect translation
0.7-0.9: Minor issues only
0.4-0.6: Significant but non-critical errors
0.1-0.3: Major errors affecting legal meaning
0.0: Completely incorrect

EVALUATION GUIDELINES:
Stylistic differences should not impact accuracy significantly unless they alter the legal meaning.
Untranslated Latin terms (e.g., prima facie) are not considered errors, but they should still be assessed for appropriate use within the context of the answer.
Terminology should be used consistently throughout the text.
Consider both explicit and implicit legal meanings.
Consider jurisdiction-specific legal terminology.
Flag any ambiguities, omissions or additions that affect legal meaning.

REQUIRED OUTPUT FORMAT:
Your response should be in plain text with the following sections:
Reasoning: Analyze how the model's translation aligns with or differs from the golden translation, focusing on significant legal and linguistic aspects.
Examples: Identify specific terms, phrases, or sections in the model's answer that were correct or incorrect, with explanations.
Score: End with exactly this format: \"The correctness score: [[score]]\"
The correctness score must strictly follow this format: \"[[score]]\", e.g., \"The correctness score: [[0.5]]\". Below are some examples.
\end{lstlisting}
\caption{The system and user prompt of the \textit{detailed} judge setup.} 
\label{tab:judge-prompt-detailed}
\end{listing*}

\begin{listing*}[h]

\textbf{System Prompt}
\begin{lstlisting}
You are a senior legal translator and quality assurance specialist with over 20 years of experience in Swiss law, certified by the Swiss Sworn Translators Association (Association suisse des traducteurs-jurés, ASTJ). You possess native-level proficiency in all Swiss national languages (German, French, Italian, and Romansh) as well as English, enabling precise evaluation of legal nuances across all linguistic combinations. Your task is to evaluate machine-translated legal texts for accuracy, clarity and fidelity to Swiss legal standards analyzing the subtle complexities of legal language. You excel at identifying even minor discrepancies and calibrating evaluation scores appropriately to reflect the severity of each error.
\end{lstlisting}

\textbf{User Prompt}
\begin{lstlisting}
GENERAL INSTRUCTIONS:
You must give each translation a score between 0 and 1 that must be divisible by 0.1 (e.g., 0.6 or 0.9). To this end, you are given a source text, its ''gold translation'' (official translation of the Swiss authorities) and the predicted translation, to which you must assign the score. You can also write down notes if deemed necessary.

SCORE: 
The scores shall reflect the completeness and accuracy of the predicted translation. In other words, you should not give a score based on readability or stylistic attributes.

POINT DEDUCTION SYSTEM:
A perfect, i.e., a perfectly complete and accurate translation receives a score of 1. 
0.1 points deduction for a relevant legal term in an unusual but still correct manner. 0.1 points shall also be deducted if the law has not been translated (e.g., BV to BV). Finally, 0.1 points shall be deducted if a non-relevant term is missing.
0.2 points deduction if a legally relevant legal term is translated erroneously. 0.2 points shall also be deducted if a relevant term is missing.
0.4 points deduction for critical errors, such as when a law is translated with reference to the wrong law.

Do not deduct points for discrepancies between the predicted translation and the gold translation if the predicted translation matches the source text better. The gold translation should primarily serve as a reference to help you assess cases where it is also a correct translation of the source. In some cases, the source text may differ slightly from the gold translation. This can happen if the source text itself was previously translated. Repeated errors for the same term should not lead to multiple point deductions. 

REQUIRED OUTPUT FORMAT:
Your response should be in plain text with the following sections:
Deductions: Focusing on significant legal and linguistic aspects, analyze and present concretely all points to be deducted together with brief explanations.
Score: End with exactly this format: \"The correctness score: [[score]]\"
The correctness score must strictly follow this format: \"[[score]]\", e.g., \"The correctness score: [[0.5]]\". Below are some examples.
\end{lstlisting}
\caption{The system and user prompt of the \textit{codebook} judge setup.} 
\label{tab:judge-prompt-codebook}
\end{listing*}

\section{Annotation Guidelines}
\label{sec:annotation-guidelines}
\begin{listing*}[h]
\small
\begin{lstlisting}
General Instructions: Annotators must give each translation a score between 0 and 10 that must be divisible by 1 (e.g., 6 or 9). To this end, annotators are given a source text, its ''gold translation'' (official translation of the Swiss authorities) and the predicted translation, to which they must assign the score. Annotators can also write down notes if deemed necessary.

Score: The scores shall reflect the completeness and accuracy of the predicted translation. In other words, annotators should not give a score based on readability or stylistic attributes.

Point Deduction System: The scoring should be conducted using a points deduction scheme.

A perfect, i.e., a perfectly complete and accurate translation receives a score of 10. 
1 points deduction for a relevant legal term in an unusual but still correct manner. 1 point shall also be deduced if the law has not been translated (e.g., BV to BV). Finally, 1 point shall be deduced if a non-relevant term is missing.
2 points deduction if a legally relevant legal term is translated erroneously. 2 points shall also be deduced if a relevant term is missing.
4 points deduction for critical errors, such as when a law is translated with reference to the wrong law. If a new category of critical error is introduced under this deduction, the annotator must inform the other annotators through their communication channel.

Do not deduct points for discrepancies between the predicted translation and the gold translation if the predicted translation matches the source text better. The gold translation should primarily serve as a reference to help you assess cases where it is also a correct translation of the source. In some cases, the source text may differ slightly from the gold translation. This can happen if the source text itself was previously translated.

Notes for Multiple Deductions: If two or more deductions are applied, annotators must briefly document the individual deductions in the comments field, e.g., ''-1, -1, -2''. This allows for potential adjustments to weighting later to account for text length if necessary. Repeated errors for the same term should not lead to multiple point deductions.

Subjectivity: We are aware that the scoring system is subject to a certain degree of subjectivity. However, assessing the quality of a translation cannot be fully objectified. To demonstrate how the scoring system works in practice, we provide annotators with 3 examples including a suggested score.

Examples:

1) Source: ''Bewilligungen nach diesem Artikel dürfen nur erteilt werden, wenn:''
Gold: ''Permits under this Article may be issued only if:''
Prediction: Permits under this Article may only be granted if:
Score: 10

2) Source: Bank client confidentiality and other client and professional confidentiality protected by law shall be maintained.
Gold: Das Bankgeheimnis und andere gesetzlich geschützte Kunden- und Berufsgeheimnisse sind zu wahren.
Prediction: Die gesetzlich geschützte Vertraulichkeit von Bankkundeninformationen sowie andere gesetzlich geschützte Kunden- und Berufsgeheimnisse sind zu wahren.
Score: 9 (-1 for unusual translation of ''Bankgeheimnis'')

3) Source: 1. La constitution de sûretés par la partie adverse (art. 79 al. 2 LBI) ne dispense pas le juge d'examiner s'il y a lieu d'ordonner des mesures provisionnelles aux conditions prévues à l'art. 77 al. 2 LBI.
Gold: 1. Eine Sicherheitsleistung gemäss Art. 79 Abs. 2 PatG enthebt den Richter nicht von der Prüfung der Frage, ob die Voraussetzungen für vorsorgliche Massnahmen nach Art. 77 Abs. 2 PatG gegeben seien.
Prediction: Die Stellung von Sicherheiten durch die Gegenpartei (Art. 79 Abs. 2 BEHG) entbindet den Richter nicht von der Prüfung, ob vorsorgliche Massnahmen unter den in Art. 77 Abs. 2 BEHG vorgesehenen Bedingungen anzuordnen sind.
Score: 6 (-4 for highly relevant erroneous translation of ''LBI'' to ''BEHG'' instead of ''PatG'')
\end{lstlisting}
\caption{The annotation guidelines given to the human experts.} 
\label{tab:annotation-guidelines}
\end{listing*}

\end{document}